\documentclass[10pt,a4paper,logo]{paper}
\usepackage[authoryear, round]{natbib}
\usepackage{wrapfig}
\usepackage{subcaption}
\usepackage{algorithm}
\usepackage{algpseudocode}
\usepackage{multirow}
\usepackage{placeins}
\usepackage{arydshln}

\makeatletter
\@namedef{ver@everyshi.sty}{}
\makeatother

\definecolor{title_blue}{HTML}{204899}
\definecolor{cite_blue}{HTML}{044dc1}
\definecolor{cite_purple}{HTML}{7406a7}
\hypersetup{
    colorlinks = true,
    citecolor = {cite_blue},
    linkcolor = {cite_purple},
    urlcolor = {cite_purple},
}

\usepackage{amsmath,amsfonts,bm}

\def\eqref#1{equation~\ref{#1}}

\def\1{\bm{1}}

\DeclareMathAlphabet{\mathsfit}{\encodingdefault}{\sfdefault}{m}{sl}
\SetMathAlphabet{\mathsfit}{bold}{\encodingdefault}{\sfdefault}{bx}{n}

\definecolor{line-blue}{RGB}{226,237,251}
\definecolor{block-grey}{gray}{0.92}
\definecolor{check-blue}{RGB}{0,102,204}
\newcommand{\cmark}{{\color{check-blue}\ding{51}}}
\newcommand{\xmark}{{\color{gray}\ding{55}}}
\newcommand{\fl}[1]{{\tiny\color{gray}($\times$#1)}}
\newcommand{\flb}[1]{{\tiny\textbf{($\times$#1)}}}
\newcommand{\ret}[2]{{\tiny\color{gray}$r{=}#1\%$ ($\times$#2)}}
\newcommand{\pc}[1]{{\tiny\color{gray}($-$#1\%)}}

\makeatletter
\renewcommand\footnoterule{%
  \kern 15\p@
  \hrule \@width 2in \kern 2.6\p@
  \vspace{4pt}
}
\makeatother

\title{Look Only When Needed: Certificate-Guided Visual Routing for Efficient Diffusion VLMs}

\author[1,\footnotesize{\textbf{*}}]{Donghyeon Cho}
\author[1,\footnotesize{\textbf{*}}]{Joowon Kim}
\author[1]{Seungho Shin}
\author[1,2,\footnotesize{$\dagger$}]{Eunho Yang}

\affil[1]{KAIST}
\affil[2]{AITRICS}

\correspondingauthor{Emails: \texttt{\{hyeon9698, kjwispro, ssh9918, eunhoy\}@kaist.ac.kr}}
\reportnumber{}

\begin{abstract}
Diffusion vision-language models (DVLMs) process their entire input at every denoising step. Since visual tokens make up 89--97\% of this input on the benchmarks we study, most of the decoding computation is spent repeatedly processing the same image. Visual token pruning methods partially relieve this burden by removing part of the visual tokens, but they lose the removed information and still process the remaining image at every step. To measure how often the image is actually needed, we compare LLaDA-V's predictions with and without the image at every decoding step and find that removing the image changes a token being unmasked at only 22.8\% of steps on average and at as few as 7.9\%. We therefore propose LOWN, a decoding method that decides at every step whether to process the image and how many tokens to unmask. LOWN predicts the masked tokens without the image and processes the image only when this prediction disagrees with the last image-conditioned prediction. Agreement certifies the earlier prediction for reuse, and the number of certified tokens sets how many are unmasked in that step. We further observe that the need for the image differs across benchmarks, so LOWN's adaptive decision is needed rather than a fixed schedule. Without training or a tuned threshold, LOWN retains 99.6\% of LLaDA-V's score on ten benchmarks at 35.3\% of the decoding FLOPs on average. LOWN also generalizes across LLaDA-based DVLMs, composes orthogonally with visual token pruning, and runs under key-value caching for further efficiency gains. Code and project page are available at \url{https://hyeon9698.github.io/LOWN/}
\end{abstract}

\begin{document}

\maketitle

\begin{figure}[t]
    \centering
    \begin{subfigure}[b]{0.69\linewidth}
        \centering
        \includegraphics[width=0.92\linewidth]{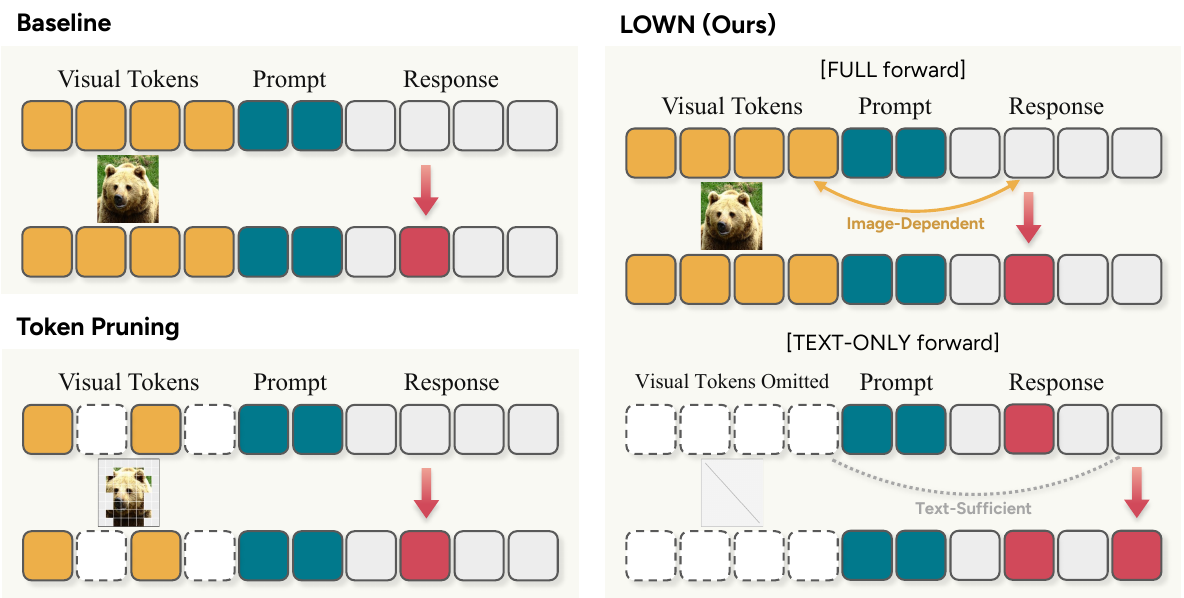}
        \caption{Overview of decoding methods for diffusion VLMs.}
        \label{fig:concept-a}
    \end{subfigure}\hspace{0.01\linewidth}%
    \begin{subfigure}[b]{0.28\linewidth}
        \centering
        \includegraphics[width=0.92\linewidth]{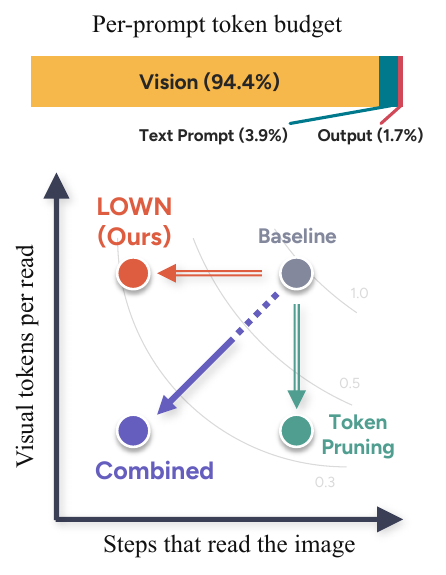}
        \captionsetup{margin={-0.06in,-0.0in}}
        \caption{Token budget and cost axes.}
        \label{fig:concept-b}
    \end{subfigure}\hspace*{0.015\linewidth}%
    \\[0.2em]
    \begin{subfigure}[b]{0.665\linewidth}
        \centering
        \includegraphics[width=0.92\linewidth]{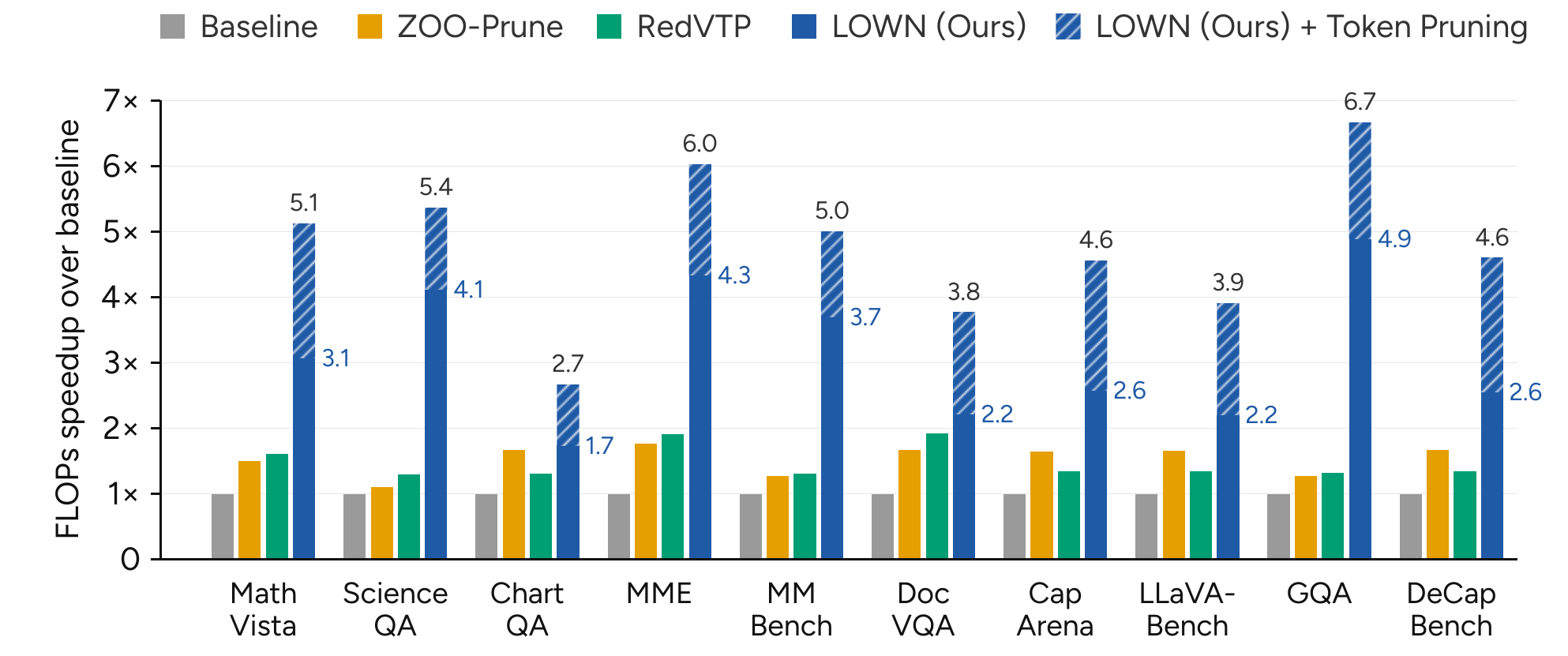}
        \caption{Relative cost at matched scores.}
        \label{fig:concept-c}
    \end{subfigure}%
    \begin{subfigure}[b]{0.32\linewidth}
        \centering
        \includegraphics[width=0.92\linewidth]{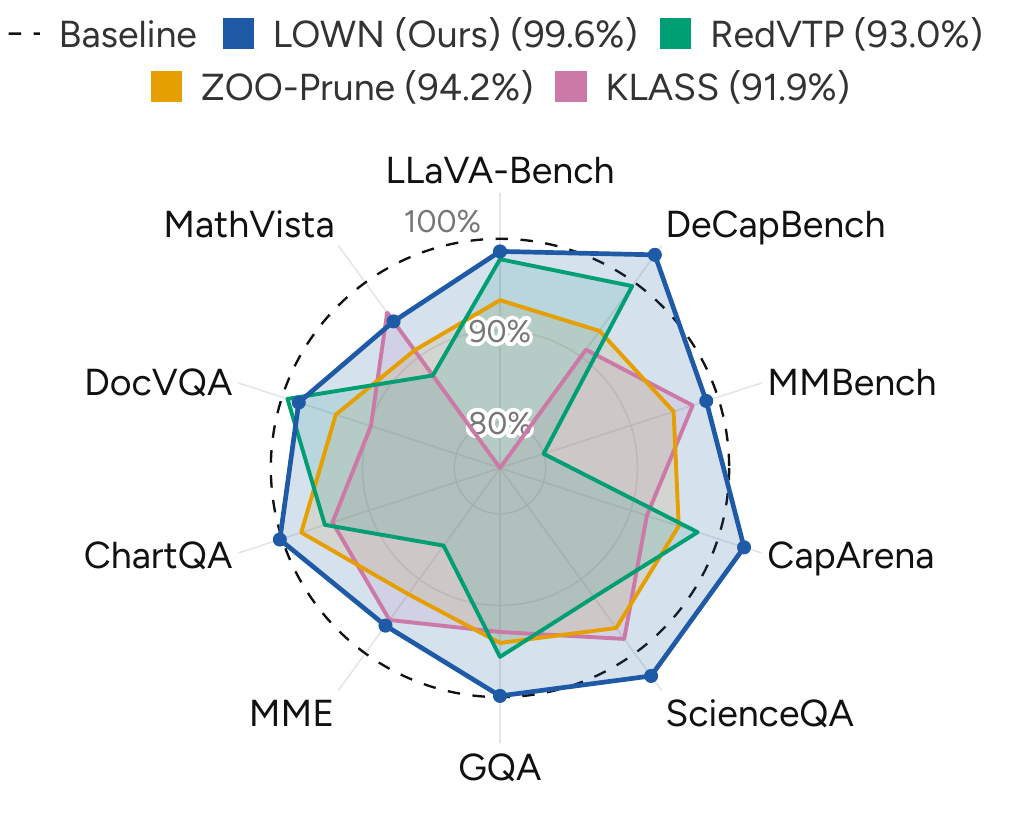}
        \caption{Relative performance at matched FLOPs.}
        \label{fig:concept-d}
    \end{subfigure}\hspace*{0.015\linewidth}%
    \caption{\textbf{Look Only When Needed (LOWN).}
    (a)~Decoding with a diffusion VLM. Token pruning reduces the visual tokens processed during decoding, whereas LOWN decides at each step whether the visual tokens need to be processed at all and otherwise runs a TEXT-ONLY forward.
    (b)~Composition of the input, averaged over ten benchmarks. Pruning reduces how many visual tokens each forward processes, LOWN whether a forward processes them at all.
    (c)~Decoding FLOPs speedup. Each pruning method is shown at LOWN's score (Appendix~\ref{app:baselines}), and LOWN + Token Pruning uses RedVTP ($r{=}50\%$).
    (d)~Quality relative to LLaDA-V when every method is run at the operating point closest to LOWN's decoding FLOPs from above. LOWN keeps 99.6\% on average.}
    \label{fig:concept}
\end{figure}

\Needspace{6\baselineskip}
\section{Introduction}
\label{sec:intro}

Recent diffusion language models match similarly sized autoregressive models while refining several masked positions in parallel over denoising steps \citep{nie2025llada,ye2025dream}. In diffusion vision-language models (DVLMs) such as LLaDA-V \citep{you2025lladav} and others \citep{yang2025mmada,yu2025dimple}, bidirectional attention prevents direct use of the exact key-value caching found in autoregressive models, so standard inference processes the full multimodal sequence at every step. Visual tokens dominate this sequence, accounting for 89--97\% of LLaDA-V's input across our benchmarks (Figure~\ref{fig:concept-b}), while a forward without them costs only 3--10\% as much. DVLM decoding therefore spends most of its compute repeatedly processing an unchanged visual input.

Existing methods reduce this cost through visual token pruning \citep{chang2025d3tom,xu2025redvtp,kim2025zooprune} (Figure~\ref{fig:concept-a}), intermediate-state caching \citep{wu2025fastdllm,liu2025dllmcache,nguyen2025elasticcache}, or unmasking more tokens per step \citep{kim2025klass,li2025prophet,wu2026coradiff,wang2026tacg,lee2026vig}. These approaches make image-conditioned forwards cheaper, reuse their computation, or run fewer decoding steps, but none asks whether the current step needs the image. We ask whether removing the image would change the model's decision at the current decoding state.

Comparing LLaDA-V's predictions with and without the image along the same decoding trajectory reveals substantial redundancy (Section~\ref{sec:observation}). The two give the same top-1 token at 83.4\% of the positions selected for unmasking, and 85.0\% of final tokens had already appeared as top-1 predictions before unmasking. Within a trajectory, image-dependent steps tend to cluster early rather than recur periodically, while their overall share varies from 7.9\% on MMBench to 43.8\% on DocVQA. A fixed schedule therefore finds it hard to capture when the image is needed. These observations motivate deciding image access per input and step from predictions the model has already computed.

We propose LOWN (Look Only When Needed), a decoding method built on this redundancy. LOWN stores the top-1 predictions and confidences from the latest image-conditioned forward, which we call the FULL forward, as an anchor and tests it with a cheap TEXT-ONLY forward at each step. Agreement certifies the anchor prediction for reuse, while disagreement triggers a new FULL forward. The confidence-ordered run of agreements also determines how many tokens to unmask. LOWN requires neither training nor a task-specific threshold and can be added to visual token pruning or key-value caching because it reduces how often a FULL forward is needed (Figure~\ref{fig:concept-b}).

We evaluate LOWN on ten benchmarks spanning multiple-choice, document and chart understanding, mathematical reasoning, and open-ended captioning. On LLaDA-V, it retains 99.6\% of the baseline score at 0.35$\times$ the decoding FLOPs (Figures~\ref{fig:concept-c} and~\ref{fig:concept-d}). Against accelerated methods given at least its FLOPs budget, it scores highest on eight of the ten benchmarks. Combined with key-value caching or visual token pruning, it further cuts the partner's FLOPs by 1.8--2.5$\times$, and with the key-value cache the pair decodes 5.6--7.0$\times$ faster than the model alone. The same rule transfers to LaViDa and VidLaDA.

Our contributions are as follows.
\begin{itemize}
    \item We show through paired FULL and TEXT-ONLY forward comparisons that most DVLM decoding steps can skip image access, while the necessary steps vary across trajectories and are not well captured by fixed schedules.
    \item We introduce LOWN, a training-free and threshold-free method that uses prediction agreement to jointly decide when to process the image and how many tokens to unmask.
    \item We validate LOWN across ten benchmarks and three DVLM backbones, retaining 99.6\% of LLaDA-V's score at 0.35$\times$ its decoding FLOPs, composing orthogonally with visual token pruning, and running under key-value caching.
\end{itemize}

\Needspace{6\baselineskip}
\section{Related Work}
\label{sec:related}

\paragraph{Diffusion language and vision-language models.}
Discrete diffusion and masked diffusion models generate sequences through iterative token recovery \citep{austin2021d3pm,sahoo2024mdlm,shi2024md4}. This paradigm has scaled to large language models \citep{nie2025llada,ye2025dream} and extended to image and video understanding \citep{diffusiongemma2026,you2025lladav,li2025lavida,yu2025dimple,he2026vidlada}. These models establish the DVLM setting studied here. LOWN addresses the complementary inference question of how often denoising must reprocess the visual input.

\paragraph{Efficient inference for diffusion LMs and VLMs.}
Visual-token reduction shortens the visual sequence in vision transformers and autoregressive VLMs \citep{bolya2023tome,chen2024fastv,zhang2025sparsevlm,xing2025pyramiddrop,shang2025prumerge,yang2025visionzip}. ZOO-Prune \citep{kim2025zooprune} combines projector-level sensitivity estimates with diversity-aware selection.

For diffusion VLMs, RedVTP prunes using attention from still-masked response tokens, while D$^3$ToM uses previously decoded tokens to guide visual-token selection and merging \citep{xu2025redvtp,chang2025d3tom}. These methods determine which visual tokens a forward processes, but they lose the removed information \citep{endo2025feather,li2026dmllmredundancy}.

Other work keeps the visual input and reduces the number of steps or the cost of a forward. Decoding schedules unmask multiple tokens per step when confidence, cross-step stability, or attention to the image supports it \citep{wei2025slowfast,wu2025fastdllm,yu2025dimple,wu2026coradiff,wang2026tacg,lee2026vig}. Prophet \citep{li2025prophet} unmasks all remaining tokens once the average top-two logit gap in the answer region is large enough. KLASS \citep{kim2025klass} combines confidence with cross-step distribution stability and is our schedule baseline. Caching methods reuse intermediate states across denoising steps \citep{ma2025dkvcache,liu2025dllmcache,nguyen2025elasticcache,li2025lavida}, and we compose LOWN with Fast-dLLM \citep{wu2025fastdllm}, a blockwise approximate key-value cache. Pruning and caching act inside a forward, on the tokens it processes or the states it reuses, and schedules on how many tokens a step unmasks. LOWN decides both whether a step processes the image and how many tokens it unmasks from the agreement between a TEXT-ONLY and the last image-conditioned prediction, and, keeping only predictions, it runs on top of pruning and caching.

\Needspace{6\baselineskip}
\section{Background and Motivation}
\label{sec:prelim}

\subsection{Preliminaries}
\label{sec:background}

Starting from a fully masked response, a DVLM repeatedly predicts response tokens over $T$ denoising steps while conditioning on fixed visual and prompt tokens \citep{sahoo2024mdlm,nie2025llada,you2025lladav}. Let $v$ denote the visual tokens, $c$ the prompt tokens, $y$ the response, $x_t$ the partially masked response at step $t$, and $M_t$ its masked positions. A \emph{FULL forward} over $(v,c,x_t)$ gives the top-1 prediction and confidence at each $i\in M_t$,
\begin{equation}
\hat y^{\mathrm{full}}_{t,i}=\arg\max_{w}\,p_\theta(y^i{=}w\mid v,c,x_t),\qquad
\hat s^{\mathrm{full}}_{t,i}=\max_{w}\,p_\theta(y^i{=}w\mid v,c,x_t).
\label{eq:full}
\end{equation}
LLaDA-V runs a FULL forward at every step and unmasks the $k$ most confident positions \citep{nie2025llada,you2025lladav}, where $k$ is the response length divided by $T$.

\subsection{When Does a DVLM Need the Image?}
\label{sec:observation}

To measure how often the DVLM uses the image, we compare two forward passes along LLaDA-V's standard decoding trajectory on all ten benchmarks of Section~\ref{sec:experiments}. The FULL forward evaluates $p_\theta(\cdot\mid v,c,x_t)$, whereas the \emph{TEXT-ONLY forward} removes the visual tokens while preserving the position IDs and evaluates $p_\theta(\cdot\mid c,x_t)$. At every $i\in M_t$, we record the top-1 predictions from both forwards, $\hat y^{\mathrm{full}}_{t,i}$ and $\hat y^{\mathrm{text}}_{t,i}$, together with the FULL confidence $\hat s^{\mathrm{full}}_{t,i}$. The FULL forward alone determines the decoding trajectory, and the TEXT-ONLY outputs are recorded without affecting its decisions. Appendix~\ref{app:replay} gives the full protocol and per-benchmark values.
\textbf{The image rarely changes a decision.} Of the positions the model unmasks, 83.4\% receive the same top-1 token from the FULL and the TEXT-ONLY forward, averaged over the ten benchmarks, so at most steps the prediction follows from the text alone and the presence of the visual tokens rarely changes the token being unmasked. Moreover, 85.0\% of the final tokens had already appeared as the top-1 prediction at the same position before the step at which they were unmasked. The baseline nevertheless continued to recompute those positions with the image until unmasking them.

\begin{figure}[t]
    \centering
    \begin{subfigure}[b]{0.775\linewidth}
        \centering
        \includegraphics[width=\linewidth,trim=40 0 0 0,clip]{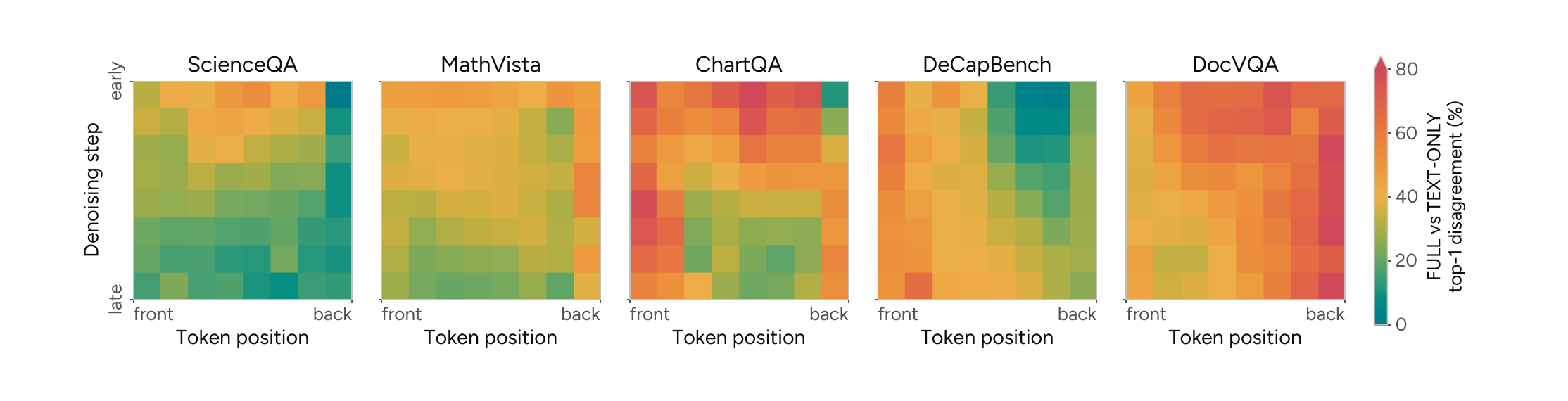}
        \caption{Where and when the image changes a prediction.}
        \label{fig:visual-dependence-a}
    \end{subfigure}\hspace{-1.5em}
    \begin{subfigure}[b]{0.245\linewidth}
        \centering
        \includegraphics[width=\linewidth,trim=0 0 12 0,clip]{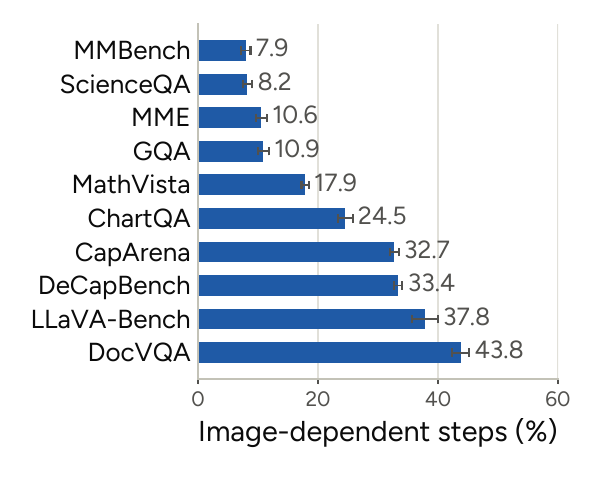}
        \captionsetup{margin=0pt}
        \caption{Image-dependent steps.}
        \label{fig:visual-dependence-b}
    \end{subfigure}
    \caption{\textbf{Visual dependence along the decoding trajectory of LLaDA-V.} (a)~Share of masked positions at which the FULL and TEXT-ONLY top-1 predictions disagree, over token position (front to back) and denoising step (early to late). (b)~Share of steps on which the image changed at least one unmasked decision, per benchmark, with 95\% confidence intervals.}
    \label{fig:visual-dependence}
\end{figure}

\textbf{The need for image access varies along the decoding trajectory.} On eight of the ten benchmarks, disagreement between the two forwards is largest in the early denoising steps and decreases as the response takes shape. This pattern is not universal, as illustrated by DeCapBench in Figure~\ref{fig:visual-dependence-a}. The affected token positions also differ across tasks, and the share of image-dependent steps ranges from 7.9\% on MMBench to 43.8\% on DocVQA, 22.8\% on average (Figure~\ref{fig:visual-dependence-b}). Variation across steps, positions, and benchmarks therefore makes a fixed image-access schedule poorly matched to the observed trajectories and motivates input- and step-adaptive image access.

A method that skips the image therefore needs a test that runs at every step, and the TEXT-ONLY forward that exposed the redundancy is a natural candidate, since it costs 3--10\% of a FULL forward. The next section turns the agreement between the two forwards into that test.

\Needspace{6\baselineskip}
\section{Method}
\label{sec:method}

\begin{figure}[t]
    \centering
    \includegraphics[width=0.9\linewidth]{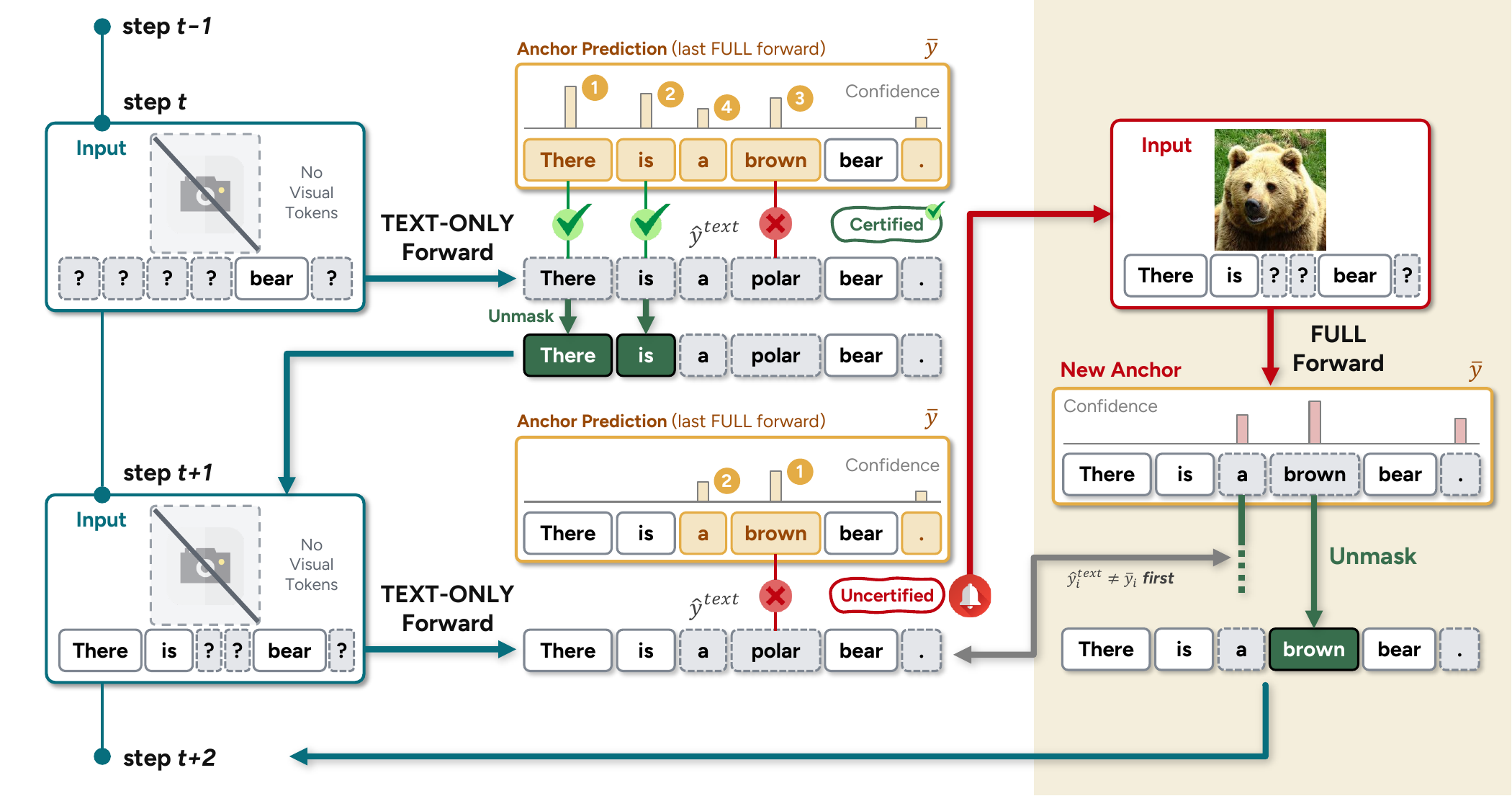}
    \caption{\textbf{Overview of LOWN.} TEXT-ONLY agreement certifies anchor tokens for image-free unmasking, and an insufficient certified prefix triggers a FULL forward that refreshes the anchor and unmasks the positions where the two forwards disagree first.}
    \label{fig:method}
\end{figure}

LOWN separates two decisions that standard DVLM decoding makes with a single FULL forward, whether the current response state needs the image and how far decoding can advance (Figure~\ref{fig:method}). It retains the last image-conditioned prediction as an anchor, tests that anchor with a TEXT-ONLY forward, and runs a FULL forward only when the test cannot certify enough tokens. The same test controls how many tokens are unmasked, allowing a successful TEXT-ONLY step to advance farther than the baseline. LOWN changes neither the model parameters nor the attention pattern.

\subsection{Anchor and Certificate}
\label{sec:setup}
\label{sec:certificate}

Whenever LOWN runs a FULL forward, it retains its predictions $\bar y$ and confidences $\bar s$ on the positions that remain masked, and we call this pair the \emph{anchor}. At step $t$, LOWN runs a TEXT-ONLY forward and orders the masked positions by decreasing anchor confidence $\bar s$. A candidate token $i$ is \emph{certified} when its current TEXT-ONLY prediction agrees with the anchor, $\hat y^{\mathrm{text}}_{t,i}=\bar y_i$. The two predictions differ in whether they see the image and in the response state they were computed from, so their agreement approximates two conditions at once, that the image is not needed for this token and that the anchor remains valid after the state has changed, without an absolute confidence threshold. In the measurement of Section~\ref{sec:observation}, a fresh FULL forward overturns 4.1--11.0\% of the tokens certified against a one-step-old anchor, and in LOWN's own runs it overturns 2.2--4.7\% of the tokens unmasked on TEXT-ONLY steps (Section~\ref{sec:ablation}).

The longest confidence-ordered prefix of certified candidates forms $\mathcal{U}^{\mathrm{text}}$. The first disagreement ends the prefix, so the acceptance length $|\mathcal{U}^{\mathrm{text}}|$ is set by the current input and step rather than by a constant. The scan also applies a repetition guard, detailed in Appendix~\ref{app:algorithm}. The certificate is checked against a prediction that ages as the response changes, and its error grows with that age (Section~\ref{sec:ablation}). LOWN therefore replaces the entire anchor whenever it runs a FULL forward, so that every certificate is checked against the most recent FULL forward.

\subsection{Certificate-Guided Decoding}
\label{sec:step}
\label{sec:acceptance}

Each step runs a TEXT-ONLY forward and forms the certified prefix $\mathcal{U}^{\mathrm{text}}$ of Section~\ref{sec:certificate}. The length of this prefix decides whether the step also needs a FULL forward. When the prefix contains at least $k$ tokens, where $k$ is the number of tokens the baseline unmasks per step, the step skips the FULL forward and accepts the whole prefix, unmasking every certified position to its anchor prediction rather than only $k$ of them. This whole-prefix acceptance is what lets one step unmask more than $k$ tokens. When the prefix is shorter than $k$, the anchor no longer certifies enough tokens for a step, so the step runs a FULL forward, replaces the anchor with its predictions and confidences, and unmasks $k$ positions to their FULL predictions in disagreement-first order. Positions whose TEXT-ONLY and FULL predictions differ, $\hat y^{\mathrm{text}}_{t,i}\ne\hat y^{\mathrm{full}}_{t,i}$, come first, and the remaining positions follow FULL confidence. These $k$ positions form $\mathcal{U}^{\mathrm{full}}$.

The resulting state update is
\begin{equation}
x_i \leftarrow
\begin{cases}
\bar y_i, & i\in\mathcal{U}^{\mathrm{text}},\quad |\mathcal{U}^{\mathrm{text}}|\ge k,\\
\hat y^{\mathrm{full}}_{t,i}, & i\in\mathcal{U}^{\mathrm{full}},\quad |\mathcal{U}^{\mathrm{text}}|<k.
\end{cases}
\label{eq:lown-update}
\end{equation}

Both branches unmask at least $k$ tokens, so LOWN never takes more than the baseline's $T$ steps, while the TEXT-ONLY branch can take fewer. Appendix~\ref{app:algorithm} provides the detailed algorithm.

\subsection{Cost Analysis}
\label{sec:guarantee}

Let $\mathcal{C}(n)$ denote the FLOPs of one forward over $n$ tokens. The FULL and TEXT-ONLY input lengths are $N_F=|v|+|c|+|x_t|$ and $N_T=|c|+|x_t|$. If the baseline runs $T$ FULL forwards, its cost is $F_{\mathrm{base}}=T\cdot\mathcal{C}(N_F)$. LOWN runs $n_T$ TEXT-ONLY forwards and adds a FULL forward on $n_F$ of those steps, which gives $F_{\mathrm{LOWN}}=n_F\cdot\mathcal{C}(N_F)+n_T\cdot\mathcal{C}(N_T)$. Relative FLOPs are therefore
\begin{equation}
R = \frac{F_{\mathrm{LOWN}}}{F_{\mathrm{base}}}
= \frac{n_F}{T} + \frac{n_T}{T}\frac{\mathcal{C}(N_T)}{\mathcal{C}(N_F)}
= f+\sigma\rho .
\label{eq:cost}
\end{equation}
Here $f=n_F/T$ is the frequency of FULL forwards, $\sigma=n_T/T\le1$ is the relative number of steps, and $\rho=\mathcal{C}(N_T)/\mathcal{C}(N_F)$ is the unit cost of a TEXT-ONLY forward. The certificate lowers $f$, and accepting more than $k$ tokens can lower $\sigma$. The input determines $\rho$, which decreases as visual tokens occupy more of the FULL input. Appendix~\ref{app:cost} gives $\mathcal{C}$ and the per-benchmark $\rho$.

Visual token pruning changes only the FULL term. Let $\pi=\mathcal{C}(N_F^{\mathrm{pruned}})/\mathcal{C}(N_F)$ be the FULL-forward cost with the pruned visual input. Pruning alone costs $\pi$ relative to the baseline, while its composition with LOWN costs $\pi f+\sigma\rho$, and both are charged the pruning overhead once. Pruning reduces the cost of each FULL forward, whereas LOWN reduces how often one is run. Section~\ref{sec:experiments} evaluates this composition.

\begin{figure}[t]
    \centering
    \includegraphics[width=\linewidth]{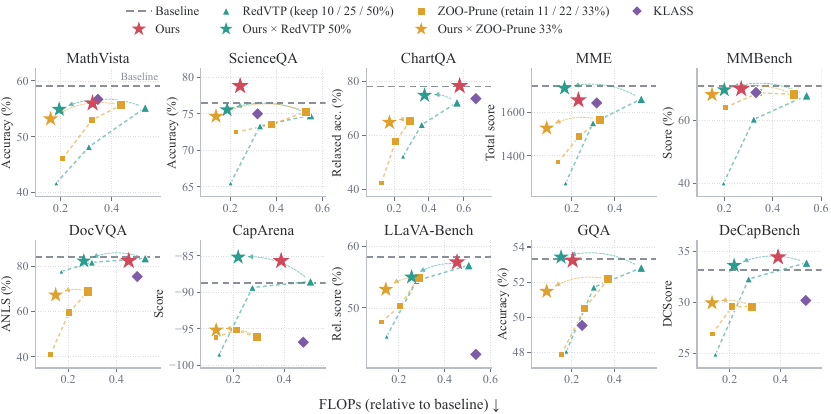}
    \caption{\textbf{Performance and efficiency on LLaDA-V}. Gray dashed lines denote the baseline. Triangles and squares denote RedVTP and ZOO-Prune retention settings, respectively. Arrows connect each fixed-pruning setting to its composition with LOWN. KLASS is matched to LOWN's FLOPs.}
    \label{fig:pareto}
\end{figure}

\Needspace{6\baselineskip}
\section{Experiments}
\label{sec:experiments}

Our experiments test whether LOWN preserves quality while reducing decoding cost. Section~\ref{sec:implementation} defines the models, benchmarks, baselines, and evaluation protocol. Section~\ref{sec:main} evaluates standalone LOWN, its composition with visual token pruning and key-value caching, and its transfer to other image and video DVLMs. Section~\ref{sec:ablation} compares image-access schedules, isolates visual routing and adaptive acceptance, and analyzes visual information preservation and certificate reliability.

\subsection{Implementation Details}
\label{sec:implementation}

\paragraph{Benchmarks and evaluation.}
We evaluate LOWN on LLaDA-V \citep{you2025lladav} and LaViDa \citep{li2025lavida} across ten benchmarks, namely MathVista \citep{lu2024mathvista}, ScienceQA \citep{lu2022scienceqa}, ChartQA \citep{masry2022chartqa}, MME \citep{fu2023mme}, MMBench \citep{liu2024mmbench}, DocVQA \citep{mathew2021docvqa}, GQA \citep{hudson2019gqa}, LLaVA-Bench \citep{liu2023llava}, CapArena \citep{cheng2025caparena}, and DeCapBench \citep{ye2025decapbench}. We additionally evaluate VidLaDA-8B \citep{he2026vidlada} on EgoSchema \citep{mangalam2023egoschema} and MLVU \citep{zhou2025mlvu}. We evaluate quality using each benchmark's task-specific metric and apply the same scorer to all methods within a comparison. Analytical decoding FLOPs are reported relative to each backbone's baseline.

\paragraph{Baselines.}
We compare RedVTP \citep{xu2025redvtp}, ZOO-Prune \citep{kim2025zooprune}, KLASS \citep{kim2025klass}, and D$^3$ToM \citep{chang2025d3tom}. All methods share prompts and response budgets within each comparison. Where multiple operating points are available, we select settings by FLOPs rather than task scores and assign each comparator at least the corresponding LOWN budget. We apply the same principle to LLaDA-V and LaViDa and evaluate Fast-dLLM \citep{wu2025fastdllm} separately in the caching comparison.

\paragraph{LOWN implementation.}
For standalone LOWN on LLaDA-V, we use batch size 1 and greedy decoding with temperature and classifier-free guidance scale set to zero, without key-value caching. The image is encoded once. FULL forwards reuse its visual embeddings, whereas TEXT-ONLY forwards remove those embeddings while preserving the original position IDs of the retained prompt and response tokens. Appendix~\ref{app:protocol} provides the complete evaluation protocol, baseline settings, and decoding details.

\subsection{Main Results}
\label{sec:main}

\begin{table}[t]
\centering
\caption{\textbf{Performance at matched decoding FLOPs on LLaDA-V}. The upper block compares standalone methods at LOWN's budget, and the lower block at the budget of LOWN with RedVTP ($r{=}50\%$). Competitors run on a discrete hyperparameter grid, so each cell uses the cheapest setting with at least LOWN's FLOPs (Appendix~\ref{app:baselines}). Cells show scores with FLOPs relative to LLaDA-V below. Best scores are bold, second best underlined, and lowest FLOPs bold.}
\label{tab:main}
\scriptsize
\setlength{\tabcolsep}{2pt}
\renewcommand{\arraystretch}{1.0}
\setlength{\aboverulesep}{0pt}\setlength{\belowrulesep}{0pt}\setlength{\extrarowheight}{1pt}
\resizebox{\linewidth}{!}{%
\begin{tabular}{l cccccccccc|c}
\toprule
 & Math & Science & Chart & & MM & Doc & Cap & LLaVA- & & DeCap & \\
\multirow{-2}{*}{Method} & Vista & QA & QA & \multirow{-2}{*}{MME} & Bench & VQA & Arena & Bench & \multirow{-2}{*}{GQA} & Bench & \multirow{-2}{*}{Avg.} \\
\midrule
LLaDA-V \textit{(Vanilla)} & 59.10 & 76.50 & 78.16 & 1719.49 & 70.88 & 83.91 & $-$88.67 & 58.20 & 53.31 & 33.18 & 100.0\% \\
\midrule
\multicolumn{11}{>{\columncolor{block-grey}}c|}{\textit{FLOPs matched to LOWN}} & \cellcolor{block-grey} \\
\midrule
\rowcolor{white} & 51.70 & 70.70 & 74.32 & 1469.78 & 56.70 & \textbf{83.39} & \underline{$-$91.00} & \underline{56.90} & \underline{50.97} & \underline{33.03} & 93.0\% \\
\multirow{-2}{*}{RedVTP} & \fl{0.354} & \fl{0.286} & \fl{0.641} & \fl{0.256} & \fl{0.280} & \fl{0.520} & \fl{0.411} & \fl{0.505} & \fl{0.260} & \fl{0.411} & \fl{0.392} \\
 & 53.70 & 73.87 & \underline{76.44} & 1582.41 & 67.27 & 78.73 & $-$93.17 & 54.30 & 50.16 & 31.02 & \underline{94.2\%} \\
\multirow{-2}{*}{ZOO-Prune} & \fl{0.403} & \fl{0.348} & \fl{0.641} & \fl{0.299} & \fl{0.317} & \fl{0.483} & \fl{0.417} & \fl{0.490} & \fl{0.238} & \fl{0.411} & \fl{0.405} \\
 & \textbf{56.70} & \underline{75.01} & 73.64 & \underline{1641.99} & \underline{68.81} & 75.36 & $-$96.83 & 42.50 & 49.52 & 30.18 & 91.9\% \\
\multirow{-2}{*}{KLASS} & \fl{0.346} & \fl{0.318} & \fl{0.669} & \fl{0.316} & \fl{0.331} & \fl{0.487} & \fl{0.476} & \fl{0.536} & \fl{0.249} & \fl{0.500} & \fl{0.423} \\
\rowcolor{line-blue} & \underline{56.00} & \textbf{78.83} & \textbf{78.36} & \textbf{1655.08} & \textbf{69.93} & \underline{82.31} & \textbf{$-$85.67} & \textbf{57.40} & \textbf{53.24} & \textbf{34.43} & \textbf{99.6\%} \\
\rowcolor{line-blue} \multirow{-2}{*}{LOWN (Ours)} & \flb{0.325} & \flb{0.243} & \flb{0.576} & \flb{0.231} & \flb{0.270} & \flb{0.452} & \flb{0.388} & \flb{0.454} & \flb{0.205} & \flb{0.391} & \flb{0.353} \\
\midrule
\multicolumn{11}{>{\columncolor{block-grey}}c|}{\textit{FLOPs matched to LOWN (Ours) + RedVTP ($r{=}50\%$)}} & \cellcolor{block-grey} \\
\midrule
\rowcolor{white} & 48.40 & 65.49 & 66.00 & 1275.08 & 56.70 & \underline{81.62} & \underline{$-$91.33} & 54.40 & 48.05 & \underline{31.16} & 87.8\% \\
\multirow{-2}{*}{RedVTP} & \fl{0.268} & \fl{0.202} & \fl{0.400} & \fl{0.171} & \fl{0.280} & \fl{0.299} & \fl{0.232} & \fl{0.278} & \fl{0.175} & \fl{0.231} & \fl{0.253} \\
 & 46.10 & 72.48 & \underline{71.68} & 1487.83 & 64.09 & 68.82 & $-$95.50 & \underline{54.90} & 47.88 & 29.12 & 88.8\% \\
\multirow{-2}{*}{ZOO-Prune} & \fl{0.208} & \fl{0.224} & \fl{0.418} & \fl{0.214} & \fl{0.205} & \fl{0.282} & \fl{0.268} & \fl{0.287} & \fl{0.152} & \fl{0.265} & \fl{0.252} \\
KLASS & \underline{52.70} & \underline{73.48} & 68.92 & \underline{1584.56} & \underline{65.29} & 76.65 & $-$96.67 & 42.50 & \underline{48.93} & 29.02 & \underline{89.3\%} \\
\quad + RedVTP \textit{($r{=}50\%$)} & \fl{0.207} & \fl{0.210} & \fl{0.428} & \fl{0.191} & \fl{0.207} & \fl{0.274} & \fl{0.247} & \fl{0.290} & \fl{0.154} & \fl{0.258} & \fl{0.247} \\
\rowcolor{line-blue} LOWN (Ours) & \textbf{54.90} & \textbf{75.56} & \textbf{74.88} & \textbf{1709.75} & \textbf{69.67} & \textbf{82.24} & \textbf{$-$85.17} & \textbf{55.00} & \textbf{53.42} & \textbf{33.61} & \textbf{98.3\%} \\
\rowcolor{line-blue} \quad + RedVTP \textit{($r{=}50\%$)} & \flb{0.195} & \flb{0.186} & \flb{0.375} & \flb{0.166} & \flb{0.200} & \flb{0.265} & \flb{0.219} & \flb{0.256} & \flb{0.150} & \flb{0.218} & \flb{0.223} \\
\bottomrule
\end{tabular}
}
\end{table}

\paragraph{Results on LLaDA-V.}
The upper block of Table~\ref{tab:main} reports LOWN's standalone results. LOWN retains 99.6\% of the baseline score on average at 0.353$\times$ the baseline FLOPs. It scores highest among the accelerated methods on eight of ten benchmarks. LOWN uses one adaptive rule without additional tuning, whereas KLASS reports benchmark-specific thresholds. The 99.6\% average includes four benchmarks on which LOWN scores above the baseline, and capping each benchmark at 100\% so that these gains do not offset losses elsewhere gives 98.6\%. Visual routing alone exceeds the baseline on ScienceQA and CapArena, and disagreement-first order lifts ChartQA and DeCapBench above it, raising the score on seven of the ten benchmarks (Table~\ref{tab:fullorder}). On a FULL step, this order first unmasks the positions where the image changed the prediction, so image-dependent tokens are not deferred.

\paragraph{Composition with visual token pruning.}
We combine LOWN with RedVTP and ZOO-Prune without retraining or changing their retention ratios. With RedVTP at 50\% retention, LOWN reaches a normalized average of 98.3 at 0.223$\times$ the baseline FLOPs and scores highest on all ten benchmarks (Table~\ref{tab:main}). It further reduces FLOPs by 57.6\% for RedVTP at 50\% retention and 60.6\% for ZOO-Prune at 33\% retention (Table~\ref{tab:pruning-relative}). For a visual-token pruner, further cost reductions require discarding more visual information. LOWN instead reduces how often the retained visual tokens are processed, allowing less aggressive pruning to outperform a more aggressive setting in both quality and compute. On MathVista, RedVTP at 50\% retention with LOWN scores 54.9 at 0.195$\times$ FLOPs, against 51.7 at 0.354$\times$ for RedVTP at 30\% retention. Figure~\ref{fig:pareto} shows the broader trade-off.

\begin{table}[h]
\centering
\caption{Composition of LOWN and Fast-dLLM on LLaDA-V.}
\label{tab:fastdllm}
\small
\setlength{\tabcolsep}{3pt}
\renewcommand{\arraystretch}{1.2}
\resizebox{0.9\linewidth}{!}{%
\begin{tabular}{@{}l c r@{\,}l r@{\,}l c r@{\,}l r@{\,}l c r@{\,}l r@{\,}l@{}}
\toprule
\multirow{2}{*}{Method} & \multicolumn{5}{c}{MathVista} & \multicolumn{5}{c}{LLaVA-Bench} & \multicolumn{5}{c}{DeCapBench} \\
\cmidrule(lr){2-6} \cmidrule(lr){7-11} \cmidrule(lr){12-16}
 & Score $\uparrow$ & \multicolumn{2}{c}{Time (s) $\downarrow$} & \multicolumn{2}{c}{FLOPs $\downarrow$} & Score $\uparrow$ & \multicolumn{2}{c}{Time (s) $\downarrow$} & \multicolumn{2}{c}{FLOPs $\downarrow$} & Score $\uparrow$ & \multicolumn{2}{c}{Time (s) $\downarrow$} & \multicolumn{2}{c}{FLOPs $\downarrow$} \\
\midrule
LLaDA-V & 59.10 & 7.56 & & 1.0000 & & 58.20 & 6.99 & & 1.0000 & & 33.18 & 14.64 & & 1.0000 & \\
\hspace{0.7em}Fast-dLLM & 57.30 & 2.14 & & 0.0856 & & 63.00 & 2.02 & & 0.0876 & & 35.19 & 4.22 & & 0.1104 & \\
\hspace{0.7em}Fast-dLLM + LOWN & 57.10 & 1.23 & {\tiny\color{check-blue}($-$42.4\%)} & 0.0457 & {\tiny\color{check-blue}($-$46.6\%)} & 58.00 & 1.25 & {\tiny\color{check-blue}($-$38.1\%)} & 0.0489 & {\tiny\color{check-blue}($-$44.1\%)} & 33.35 & 2.10 & {\tiny\color{check-blue}($-$50.1\%)} & 0.0453 & {\tiny\color{check-blue}($-$58.9\%)} \\
\bottomrule
\end{tabular}}
\end{table}

\paragraph{Composition with key-value caching.}
LOWN also provides plug-and-play acceleration on top of key-value caching, since the cache lowers the cost of each forward and LOWN's acceptance lowers the number of forwards. We retain Fast-dLLM's cache block and refresh settings and combine its token selection with LOWN's acceptance rule. Across MathVista, LLaVA-Bench, and DeCapBench, adding LOWN reduces mean generation time by 38.1--50.1\% (1.61--2.01$\times$ further speedup) and FLOPs by 44.1--58.9\% relative to Fast-dLLM (Table~\ref{tab:fastdllm}, Appendix~\ref{app:wallclock}). The combined scores are within 2.0 points of LLaDA-V, but lower than Fast-dLLM on all three tasks, with a decrease of 5.0 points on LLaVA-Bench.

\begin{figure}[t]
    \centering
    \includegraphics[width=\linewidth]{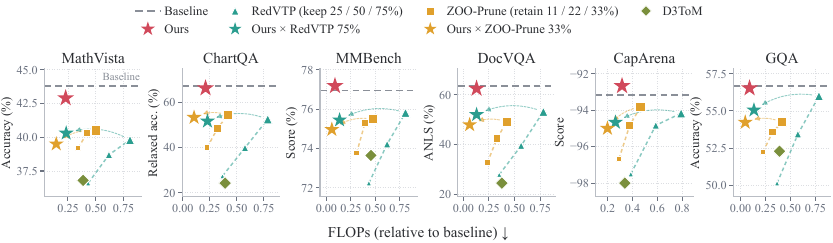}
    \caption{\textbf{Performance and efficiency on LaViDa}. Gray dashed lines denote the baseline. Triangles, squares, and diamonds denote RedVTP, ZOO-Prune, and D$^3$ToM, respectively. Composition markers use RedVTP at 75\% and ZOO-Prune at 33\% retention.}
    \label{fig:pareto-lavida}
\end{figure}

\paragraph{Other models.}
\textit{LaViDa.} LOWN transfers to LaViDa without changing its decision rule, retaining 99.7\% of the baseline score on average at 0.174$\times$ its decoding FLOPs (Figure~\ref{fig:pareto-lavida}). On MMBench and GQA, LOWN scores 77.18 versus 76.95 and 56.54 versus 56.69 at only 0.088$\times$ and 0.091$\times$ the FLOPs. As on LLaDA-V, adding LOWN to RedVTP at 75\% or ZOO-Prune at 33\% retention cuts their FLOPs by a further 77.3\% and 72.2\% at quality indices of 99.5 and 99.8. These results show that adaptive image access transfers across DVLM backbones and remains complementary to visual token pruning. Appendix~\ref{app:tables} reports the complete results.

\begin{table}[h]
\centering
\caption{Video-understanding results of LOWN on VidLaDA-8B.}
\label{tab:vidlada}
\footnotesize
\begin{tabular}{@{}l c r@{\,}l c r@{\,}l@{}}
\toprule
& \multicolumn{3}{c}{EgoSchema} & \multicolumn{3}{c}{MLVU} \\
\cmidrule(lr){2-4}\cmidrule(l){5-7}
Method & Score $\uparrow$ & \multicolumn{2}{c}{FLOPs $\downarrow$} & Score $\uparrow$ & \multicolumn{2}{c}{FLOPs $\downarrow$} \\
\midrule
VidLaDA & 60.6 & 1.000 & & 41.8 & 1.000 & \\
LOWN (Ours) & 60.2 & 0.255 & {\tiny\color{check-blue}($-$74.5\%)} & 43.3 & 0.272 & {\tiny\color{check-blue}($-$72.8\%)} \\
\bottomrule
\end{tabular}
\end{table}

\noindent\textit{VidLaDA.} We apply LOWN to VidLaDA-8B on two video benchmarks under a controlled 128-token decoding protocol (Table~\ref{tab:vidlada}, Appendix~\ref{app:video}). LOWN uses 0.255$\times$ the baseline decoding FLOPs on EgoSchema and 0.272$\times$ on MLVU, with scores of 60.2 and 43.3 compared with the baseline's 60.6 and 41.8.

\paragraph{Wall-clock latency.}
Across the ten LLaDA-V benchmarks, standalone LOWN reduces generation time by 57.4\% on average and up to 73.2\%. Appendix~\ref{app:wallclock} repeats Table~\ref{tab:main} with each competitor given at least LOWN's generation time, for standalone methods and pruning compositions.

\subsection{Ablation and Analysis}
\label{sec:ablation}

\begin{table}[h]
\centering
\caption{Comparison of LOWN with fixed and random image-access schedules on LLaDA-V. Cells report score and relative FLOPs.}
\label{tab:when}
\footnotesize
\setlength{\tabcolsep}{4pt}
\begin{tabular}{@{}l ccc@{}}
\toprule
 & MathVista & LLaVA-Bench & DeCapBench \\
\midrule
LOWN (Ours) & 56.00 \fl{0.325} & 57.40 \fl{0.454} & 34.43 \fl{0.391} \\
\midrule
Period 2 & 57.00 \fl{0.541} & 56.10 \fl{0.516} & 33.43 \fl{0.522} \\
Period 4 & 52.00 \fl{0.321} & 49.00 \fl{0.274} & 27.79 \fl{0.284} \\
Period 8 & 48.50 \fl{0.211} & 41.30 \fl{0.154} & 19.09 \fl{0.164} \\
Random & 53.40 \fl{0.349} & 53.80 \fl{0.456} & 30.48 \fl{0.389} \\
\bottomrule
\end{tabular}
\end{table}

\paragraph{Image-access scheduling.}
Table~\ref{tab:when} compares LOWN with periodic and random image access. Period $n$ uses the image every $n$ steps, while Random matches LOWN's average FULL-forward count. Against this matched random schedule, LOWN improves task-normalized quality by 4.9--13.0\% on all three tasks. Period 4 uses similar compute to LOWN on MathVista but scores 4.0 points lower.

\paragraph{Visual routing and adaptive acceptance.}
Table~\ref{tab:components} separates two decisions of LOWN on LLaDA-V. \textit{Visual routing} is the certificate's choice of the steps that run a FULL forward, and \textit{adaptive acceptance} is whole-prefix acceptance on TEXT-ONLY steps together with disagreement-first order on FULL steps (Section~\ref{sec:step}, Appendix~\ref{app:fullorder}). Visual routing alone, which keeps the baseline's order and unmasks $k$ tokens per step, reduces FLOPs by 27--68\%. Adaptive acceptance removes a further 21--43\% of the remaining cost, scores higher on seven of the ten benchmarks, and loses at most 2.8\% of normalized quality on the others. Most of the saving therefore comes from skipping FULL forwards, and adaptive acceptance adds a task-dependent reduction. This is not the same as running fewer steps, since we find that uniformly using two or four times fewer steps degrades performance across benchmarks (Table~\ref{tab:fewer-steps}).

\begin{table}[t]
\centering
\caption{Ablation of visual routing and adaptive acceptance on LLaDA-V. Each setting reports scores and, below, relative FLOPs with the reduction from the preceding setting in parentheses.}
\label{tab:components}
\scriptsize
\setlength{\tabcolsep}{2pt}
\renewcommand{\arraystretch}{1.0}
\resizebox{\linewidth}{!}{%
\begin{tabular}{@{}cc cccccccccc@{}}
\toprule
 & Adaptive & Math & Science & Chart & & MM & Doc & Cap & LLaVA- & & DeCap \\
\multirow{-2}{*}{Visual routing} & acceptance & Vista & QA & QA & \multirow{-2}{*}{MME} & Bench & VQA & Arena & Bench & \multirow{-2}{*}{GQA} & Bench \\
\midrule
\xmark & \xmark & 59.10 & 76.50 & 78.16 & 1719.5 & 70.88 & 83.91 & $-$88.67 & 58.20 & 53.31 & 33.18 \\
\addlinespace[1.5pt]\hdashline[0.5pt/1.5pt]\addlinespace[1.5pt]
 & & 57.60 & 77.39 & 77.76 & 1624.6 & 68.30 & 83.50 & $-$86.83 & 57.90 & 52.53 & 33.02 \\
\multirow{-2}{*}{\cmark} & \multirow{-2}{*}{\xmark} & 0.5040 \pc{49.6} & 0.3554 \pc{64.5} & 0.7330 \pc{26.7} & 0.3383 \pc{66.2} & 0.3759 \pc{62.4} & 0.6743 \pc{32.6} & 0.6833 \pc{31.7} & 0.7002 \pc{30.0} & 0.3203 \pc{68.0} & 0.6840 \pc{31.6} \\
\addlinespace[1.5pt]\hdashline[0.5pt/1.5pt]\addlinespace[1.5pt]
 & & 56.00 & 78.83 & 78.36 & 1655.1 & 69.93 & 82.31 & $-$85.67 & 57.40 & 53.24 & 34.43 \\
\multirow{-2}{*}{\cmark} & \multirow{-2}{*}{\cmark} & 0.3252 \pc{35.5} & 0.2428 \pc{31.7} & 0.5759 \pc{21.4} & 0.2307 \pc{31.8} & 0.2701 \pc{28.1} & 0.4517 \pc{33.0} & 0.3885 \pc{43.1} & 0.4543 \pc{35.1} & 0.2045 \pc{36.2} & 0.3909 \pc{42.9} \\
\bottomrule
\end{tabular}}
\end{table}

\begin{figure}[t]
    \centering
    \includegraphics[width=0.95\linewidth]{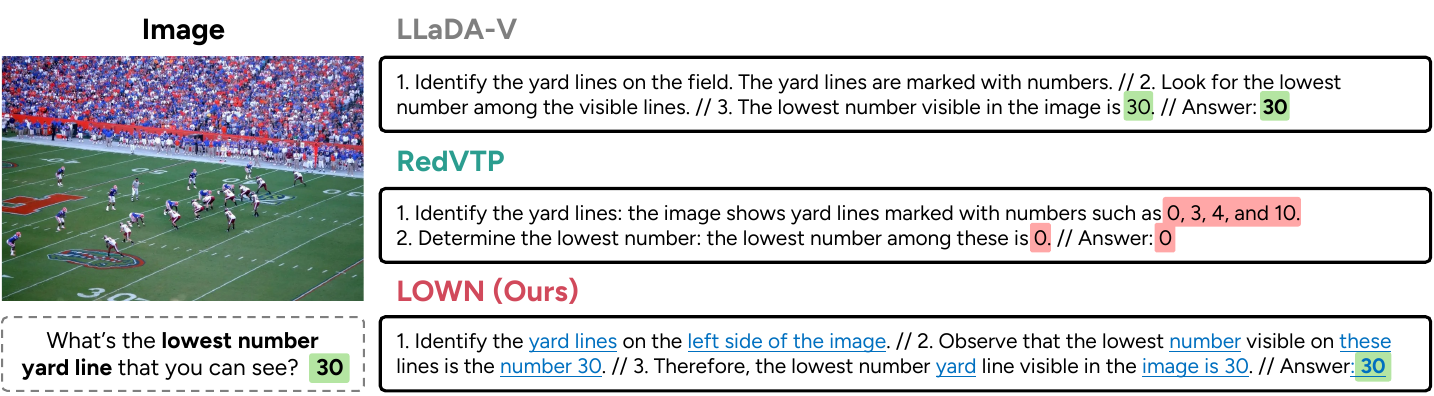}
    \caption{\textbf{Qualitative comparison with visual token pruning (RedVTP).} In LOWN's output, blue underlines mark tokens unmasked on FULL forward steps, while the remaining tokens are unmasked on TEXT-ONLY forward steps.}
    \label{fig:LOWN_vs_Exsitings}
\end{figure}

\paragraph{Visual information preservation.}
Visual token pruning permanently removes part of the visual representation before decoding, whereas LOWN skips image access only at selected steps and can consult the complete image later. The distinction is most apparent on tasks that require reading or composing fine-grained visual evidence, where pruning incurs its largest quality losses (Figure~\ref{fig:pareto-lavida} and Table~\ref{tab:lavida}). In Figure~\ref{fig:LOWN_vs_Exsitings}, LLaDA-V and LOWN use the relevant visual evidence and answer correctly, whereas RedVTP prunes that evidence and answers fluently but incorrectly. Here LOWN concentrates FULL-forward steps on visually grounded content, while surrounding language is handled by TEXT-ONLY forwards.

\paragraph{Certificate reliability.}
On the baseline decoding trajectories of Section~\ref{sec:observation}, we compare each certified token with a fresh FULL prediction. With a one-step-old anchor, the disagreement rate is 4.1--11.0\%, and it generally increases with anchor age, reaching 25.3\% on DocVQA when the anchor is eight steps old. On LOWN's own trajectories, the same rate is 2.2--4.7\% for the tokens it unmasks on TEXT-ONLY steps. In practice the anchor rarely gets old, since every FULL forward replaces it and 92--97\% of these tokens use an anchor from the preceding step, none older than five (Table~\ref{tab:anchor-age}). An oracle schedule that uses the image only on the steps where it changes an unmasked decision costs 0.33$\times$ the baseline against LOWN's measured 0.35$\times$, and, since it still runs one forward per baseline step, it is not a lower bound. Appendix~\ref{app:analysis} gives per-benchmark results.

\Needspace{6\baselineskip}
\section{Conclusion}
\label{sec:conclusion}

DVLMs need not process visual tokens at every denoising step. Along LLaDA-V's decoding trajectories, most unmasked decisions agree between FULL and TEXT-ONLY forwards, but the steps requiring visual input vary across tasks, motivating adaptive rather than fixed image access. LOWN turns this agreement into a training- and threshold-free rule for deciding when to run a FULL forward and how many tokens to unmask. Across ten benchmarks, it retains 99.6\% of LLaDA-V's score at 0.35$\times$ the decoding FLOPs. The same rule transfers unchanged to LaViDa and extends to VidLaDA under a controlled video protocol, while also composing orthogonally with visual token pruning and running under key-value caching. These results establish image-access scheduling as a broadly applicable axis for efficient DVLM decoding.

\paragraph{Limitations.}
LOWN reuses anchor predictions that agree with the TEXT-ONLY forward, so a reused token can differ from a fresh FULL prediction, more often as the anchor ages (Appendix~\ref{app:analysis}). It also reduces cost less on chart and document reading, where visual dependence persists and more steps need a FULL forward (Appendix~\ref{app:limitation}). The same agreement decides how many FULL forwards a response needs, so its cost cannot be set before decoding. Combining LOWN with visual token pruning, however, partly addresses this, as its retention ratio gives some control over cost.

\clearpage
\bibliographystyle{plainnat}
\bibliography{references}

\clearpage
\appendix

\FloatBarrier
\Needspace{6\baselineskip}
\section{Measurement Protocol and Additional Observations}
\label{app:replay}

\subsection{Measurement Protocol}
\label{app:replay-protocol}
We decode each benchmark with LLaDA-V's released procedure, greedy at temperature zero with zero classifier-free guidance and low-confidence remasking, using the response lengths and step counts of Appendix~\ref{app:protocol}. At every step we run the TEXT-ONLY forward alongside the FULL forward on the same partially masked response, with the visual tokens removed and the position IDs of all other tokens kept. Only the FULL forward decides which positions are unmasked, and the TEXT-ONLY outputs never enter the decoding state, so the trajectory is the baseline's by construction. At every masked position and step we log the top-1 token and its softmax confidence from both forwards, together with the positions the FULL forward unmasks.

\subsection{Per-Benchmark Observations}
\label{app:obs}
\begin{figure}[h]
    \centering
    \includegraphics[width=0.80\linewidth,trim=57 36 70 37,clip]{figures/fig2_v5_visual_dependence_a.pdf}\\[0.3em]
    \includegraphics[width=0.80\linewidth,trim=57 36 70 37,clip]{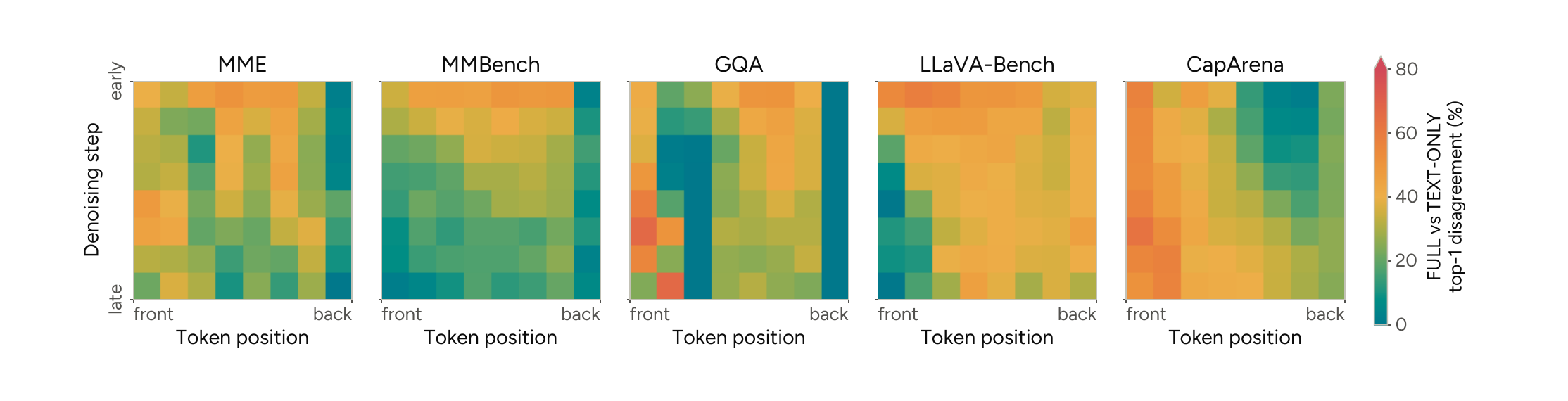}
    \captionsetup{aboveskip=3pt}
    \caption{\textbf{Where and when the image changes a prediction, all ten benchmarks.} Share of masked positions at which the FULL and TEXT-ONLY top-1 predictions disagree, over token position (front to back) and denoising step (early to late), as in Figure~\ref{fig:visual-dependence-a}.}
    \label{fig:visual-dependence-all}
\end{figure}

Table~\ref{tab:obs} lists, for each benchmark, the quantities that Section~\ref{sec:observation} and Figure~\ref{fig:concept-b} summarize. The averages in the main text are unweighted means over the ten benchmarks. The FULL and TEXT-ONLY forwards agree on 71.8--92.1\% of unmasked decisions, and 74.0--90.2\% of final tokens were already the top-1 prediction before being unmasked. The latter is lowest on the three open-ended benchmarks, LLaVA-Bench, CapArena, and DeCapBench. Figure~\ref{fig:visual-dependence-all} shows the disagreement maps of Figure~\ref{fig:visual-dependence-a} for all ten benchmarks.

\begin{table}[h]
\centering
\caption{Per-benchmark observations on LLaDA-V. Visual tokens is their mean percentage of the FULL input. Unchanged decisions is the share of unmasked positions at which the FULL and TEXT-ONLY top-1 predictions agree. Image-dependent steps is the share of steps on which the image changed at least one unmasked decision. Already top-1 is the share of final tokens that were the top-1 prediction at their position before the step that unmasked them. Brackets give 95\% confidence intervals, and the last row is the unweighted mean.}
\label{tab:obs}
\small
\setlength{\tabcolsep}{5pt}
\begin{tabular}{@{}lrrrr@{}}
\toprule
Benchmark & Visual tokens (\%) & Unchanged decisions & Image-dependent steps & Already top-1 \\
\midrule
MathVista & 89.0 & 89.0 {\scriptsize [88.6, 89.3]} & 17.9 {\scriptsize [17.2, 18.5]} & 90.2 {\scriptsize [89.9, 90.6]} \\
ScienceQA & 91.6 & 91.8 {\scriptsize [91.0, 92.5]} & 8.2 {\scriptsize [7.5, 9.0]} & 88.2 {\scriptsize [87.2, 89.0]} \\
ChartQA & 96.1 & 75.5 {\scriptsize [74.2, 76.7]} & 24.5 {\scriptsize [23.3, 25.8]} & 88.4 {\scriptsize [87.4, 89.3]} \\
MME & 97.0 & 89.4 {\scriptsize [88.5, 90.3]} & 10.6 {\scriptsize [9.7, 11.5]} & 89.6 {\scriptsize [88.7, 90.4]} \\
MMBench & 91.3 & 92.1 {\scriptsize [91.3, 92.8]} & 7.9 {\scriptsize [7.2, 8.7]} & 88.6 {\scriptsize [87.6, 89.5]} \\
DocVQA & 96.6 & 71.8 {\scriptsize [70.9, 72.7]} & 43.8 {\scriptsize [42.4, 45.2]} & 86.4 {\scriptsize [85.7, 87.1]} \\
CapArena & 94.8 & 79.6 {\scriptsize [79.1, 80.0]} & 32.7 {\scriptsize [32.0, 33.5]} & 76.9 {\scriptsize [76.5, 77.4]} \\
LLaVA-Bench & 96.2 & 76.3 {\scriptsize [74.9, 77.6]} & 37.8 {\scriptsize [35.6, 40.0]} & 74.0 {\scriptsize [72.6, 75.3]} \\
GQA & 96.5 & 89.1 {\scriptsize [88.2, 90.0]} & 10.9 {\scriptsize [10.0, 11.8]} & 89.4 {\scriptsize [88.5, 90.2]} \\
DeCapBench & 94.9 & 79.5 {\scriptsize [79.1, 79.9]} & 33.4 {\scriptsize [32.7, 34.1]} & 78.0 {\scriptsize [77.6, 78.5]} \\
\midrule
Mean & 94.4 & 83.4 & 22.8 & 85.0 \\
\bottomrule
\end{tabular}
\end{table}

\FloatBarrier
\subsection{Measurement on LaViDa}
\label{app:replay-lavida}
The measurement of Section~\ref{sec:observation} was repeated on LaViDa over the full evaluation sets of all ten benchmarks, with the response lengths and step counts of the LLaDA-V protocol so that the two models are measured on the same schedules. Table~\ref{tab:obs-lavida} lists the per-benchmark values. Each per-benchmark observation holds. On the eight question-answering benchmarks each measure shows less image dependence than on LLaDA-V, apart from already top-1 on MathVista, which is level, and on the two captioning benchmarks the two models are close. The image changes an unmasked decision on fewer steps, 1.3--30.6\% against 7.9--43.8\% on LLaDA-V, the two forwards agree on 81.6--98.7\% of unmasked decisions, and 78.6--93.7\% of final tokens were the top-1 prediction before they were unmasked.

\begin{table}[h]
\centering
\caption{Per-benchmark observations on LaViDa, in the format of Table~\ref{tab:obs}, with the response lengths and step counts of the LLaDA-V measurement and over the full evaluation sets. Brackets give 95\% confidence intervals, and the last row is the unweighted mean.}
\label{tab:obs-lavida}
\small
\setlength{\tabcolsep}{5pt}
\begin{tabular}{@{}lrrrr@{}}
\toprule
Benchmark & Visual tokens (\%) & Unchanged decisions & Image-dependent steps & Already top-1 \\
\midrule
MathVista & 71.1 & 94.0 {\scriptsize [93.8, 94.1]} & 10.4 {\scriptsize [10.1, 10.7]} & 89.8 {\scriptsize [89.6, 90.0]} \\
ScienceQA & 76.4 & 98.7 {\scriptsize [98.6, 98.8]} & 1.3 {\scriptsize [1.2, 1.4]} & 93.7 {\scriptsize [93.5, 94.0]} \\
ChartQA & 88.8 & 83.6 {\scriptsize [83.2, 84.0]} & 16.4 {\scriptsize [16.0, 16.8]} & 92.6 {\scriptsize [92.4, 92.9]} \\
MME & 87.9 & 96.1 {\scriptsize [95.9, 96.3]} & 3.9 {\scriptsize [3.7, 4.1]} & 93.7 {\scriptsize [93.5, 94.0]} \\
MMBench & 78.4 & 96.8 {\scriptsize [96.6, 96.9]} & 3.2 {\scriptsize [3.1, 3.4]} & 93.7 {\scriptsize [93.5, 93.9]} \\
DocVQA & 88.3 & 89.4 {\scriptsize [89.3, 89.6]} & 15.2 {\scriptsize [14.9, 15.4]} & 91.6 {\scriptsize [91.4, 91.7]} \\
CapArena & 81.1 & 82.0 {\scriptsize [81.7, 82.3]} & 29.8 {\scriptsize [29.3, 30.2]} & 78.6 {\scriptsize [78.3, 78.9]} \\
LLaVA-Bench & 86.3 & 91.3 {\scriptsize [90.4, 92.1]} & 13.6 {\scriptsize [12.2, 15.1]} & 88.1 {\scriptsize [87.1, 89.0]} \\
GQA & 88.8 & 95.8 {\scriptsize [95.7, 95.9]} & 4.2 {\scriptsize [4.1, 4.3]} & 93.1 {\scriptsize [92.9, 93.2]} \\
DeCapBench & 80.7 & 81.6 {\scriptsize [81.2, 81.9]} & 30.6 {\scriptsize [30.0, 31.2]} & 78.6 {\scriptsize [78.2, 79.0]} \\
\midrule
Mean & 82.8 & 90.9 & 12.9 & 89.3 \\
\bottomrule
\end{tabular}
\end{table}

\FloatBarrier
\Needspace{6\baselineskip}
\section{Method Details and Cost Accounting}
\label{app:cost}

This appendix gives the complete decoding procedure, restates the cost model of Section~\ref{sec:guarantee} and derives its forward cost $\mathcal{C}(n)$, and writes visual token pruning in the same form as Equation~\ref{eq:cost}.

\subsection{Detailed Decoding Algorithm}
\label{app:algorithm}
Each response starts fully masked, with its length and baseline unmasking schedule set by the benchmark protocol. The first iteration runs TEXT-ONLY followed by FULL on the same response, establishing the anchor and unmasking $k$ positions. Confidence is the softmax probability of the top-1 token. The repetition guard ends the certified prefix before a candidate whose token ID matches either immediate neighbor that is already unmasked or was accepted earlier in the same scan. The guard is needed because the positions accepted on one TEXT-ONLY step are predicted independently of one another, so two adjacent positions can receive the same token, and without the guard such pairs would be unmasked together and appear as repeated words in the response. Algorithm~\ref{alg:step} gives the full procedure.

\begin{algorithm}[t]
\caption{LOWN decoding}
\label{alg:step}
\begin{algorithmic}[1]
\Require visual tokens $v$, prompt tokens $c$, tokens unmasked per step $k$
\Ensure response $x$
\State $x \gets$ fully masked response;\quad $\hat y^{\mathrm{text}} \gets \arg\max p_\theta(\cdot \mid c, x)$ \Comment{draft for the first forward}
\While{$x$ has masked positions}
    \State $\bar y \gets \arg\max p_\theta(\cdot \mid v, c, x)$,\quad $\bar s \gets p_\theta(\bar y \mid v, c, x)$ \Comment{FULL forward: update anchor}
    \State $\mathcal{U}^{\mathrm{full}} \gets$ top-$k$ masked positions by $\bar s$, those with $\hat y^{\mathrm{text}}_i \neq \bar y_i$ first
    \State $x_i \gets \bar y_i$ for $i \in \mathcal{U}^{\mathrm{full}}$
    \Repeat
        \State $\hat y^{\mathrm{text}} \gets \arg\max p_\theta(\cdot \mid c, x)$ \Comment{TEXT-ONLY forward}
        \State $M \gets$ masked positions in decreasing $\bar s$
        \State $\mathcal{U}^{\mathrm{text}} \gets$ longest prefix of $M$ with $\hat y^{\mathrm{text}}_i = \bar y_i$ \Comment{certificate}
        \State \textbf{if} $|\mathcal{U}^{\mathrm{text}}| \ge k$ \textbf{then} $x_i \gets \bar y_i$ for $i \in \mathcal{U}^{\mathrm{text}}$ \Comment{anchor kept}
        \State \textbf{if} $x$ has no masked positions \textbf{then break}
    \Until{$|\mathcal{U}^{\mathrm{text}}| < k$} \Comment{a certificate fails}
\EndWhile
\State \Return $x$
\end{algorithmic}
\end{algorithm}

\subsection{Cost of One Forward}
Section~\ref{sec:guarantee} writes the cost of LOWN relative to the baseline as $R=f+\sigma\rho$ (Equation~\ref{eq:cost}). The baseline runs $T$ FULL forwards over $N_F=|v|+|c|+|x_t|$ tokens. LOWN runs $n_T$ TEXT-ONLY forwards over $N_T=|c|+|x_t|$ tokens and adds a FULL forward on $n_F$ of those steps, so $f=n_F/T$ is the frequency of FULL forwards, $\sigma=n_T/T$ the relative number of steps, and $\rho=\mathcal{C}(N_T)/\mathcal{C}(N_F)$ the cost of a TEXT-ONLY forward relative to a FULL one, where $\mathcal{C}(n)$ is the FLOPs of one forward over $n$ tokens. A FULL step is charged both forwards. This section gives $\mathcal{C}(n)$.

We count matrix multiplications only, with one multiply-add as two FLOPs. For a model with $L$ layers, hidden width $d$, FFN width $m$, and vocabulary $\mathcal{V}$ (for LLaDA-V, $L{=}32$, $d{=}4096$, $m{=}12288$, $|\mathcal{V}|{=}126{,}349$), one layer over $n$ tokens costs
\begin{equation}
\mathcal{C}_{\mathrm{layer}}(n) \;=\; \underbrace{8nd^{2} + 4n^{2}d}_{\text{attention}} \;+\; \underbrace{6ndm}_{\text{FFN}},
\qquad
\mathcal{C}(n) \;=\; L \cdot \mathcal{C}_{\mathrm{layer}}(n) + 2nd|\mathcal{V}|,
\label{eq:forward-cost}
\end{equation}
where $8nd^2$ is the Q/K/V/O projections, $4n^2 d$ is $QK^{\top}$ and $AV$ and is quadratic in length, $6ndm$ is the three matrices of the gated FFN, and the last term of $\mathcal{C}(n)$ is the LM head. The quadratic term is kept because it is what makes $\rho$ smaller than the length ratio $N_T/N_F$, since removing $v$ shrinks $n$ by roughly 9--33$\times$ and the quadratic term by roughly the square of that. The count excludes the vision encoder, which runs once per sample and would add 0.16--1.89\% of the baseline decoding cost on the ten benchmarks, the same for every method on a benchmark, and non-matmul operations such as softmax and normalization. Table~\ref{tab:rho} reports the mean $\rho$ per benchmark, and Appendix~\ref{app:wallclock} reports measured generation time.

\begin{table}[ht]
\centering
\caption{Mean TEXT-ONLY/FULL forward-cost ratio on LLaDA-V. For each sample, we compute $\rho_i=\mathcal{C}(p_i+o_i)/\mathcal{C}(p_i+v_i+o_i)$ using its prompt length $p_i$, visual-token count $v_i$, and configured response length $o_i$. Each row averages these ratios over the benchmark's samples, and $\mathcal{C}$ is the analytical forward cost in Equation~\ref{eq:forward-cost}.}
\label{tab:rho}
\small
\begin{tabular}{@{}lrr@{}}
\toprule
Benchmark & Samples & Mean $\rho$ (\%) \\
\midrule
MathVista & 1{,}000 & 10.07 \\
ScienceQA & 2{,}017 & 7.92 \\
ChartQA & 2{,}500 & 2.54 \\
MME & 2{,}374 & 3.15 \\
MMBench & 4{,}329 & 7.04 \\
DocVQA & 5{,}349 & 3.02 \\
CapArena & 600 & 4.63 \\
LLaVA-Bench & 60 & 3.35 \\
GQA & 12{,}578 & 3.20 \\
DeCapBench & 400 & 4.49 \\
\bottomrule
\end{tabular}
\end{table}

\subsection{Other Methods in the Same Form}
Relative to the unpruned baseline, visual token pruning costs $\pi$, LOWN costs $f + \sigma\rho$, and the two together cost $\pi f + \sigma\rho$. Pruning fixes its budget $\pi$ before decoding starts, whereas LOWN decides $f$ and $\sigma$ during decoding, per input and per step. Key-value caching instead lowers the cost of every forward that reuses the cache, and LOWN's acceptance lowers the number of forwards, which is the saving measured in Table~\ref{tab:fastdllm}.

\FloatBarrier
\Needspace{6\baselineskip}
\section{Experimental Setup}
\label{app:protocol}

\subsection{Benchmarks and Settings}
Table~\ref{tab:rho} lists the evaluated sample counts for LLaDA-V. We use greedy decoding with temperature zero and classifier-free guidance zero, with low-confidence remasking. The configured response length and baseline denoising steps are 100/50 for MathVista, 32/16 for DocVQA, 64/32 for LLaVA-Bench, 128/64 for CapArena and DeCapBench, and 16/16 for ScienceQA, ChartQA, MME, MMBench, and GQA. ChartQA generation ends at the first newline. The baseline unmasks $k$ tokens per step, the response length divided by the number of steps, so $k=2$ on MathVista, DocVQA, LLaVA-Bench, CapArena, and DeCapBench and $k=1$ on the others. MathVista uses the model's reasoning prompt, and the other tasks use the non-reasoning setting. These response budgets are held fixed within each comparison. For standalone LLaDA-V, we use the authors' RedVTP code and adapt the released ZOO-Prune and KLASS implementations. The released ZOO-Prune code targets autoregressive LLaVA models. We keep its token scoring and selection, select the visual tokens once before decoding, and use the retained tokens at every denoising step. Reported FLOPs use the tokens actually retained and include the scoring cost. Composition comparisons target the corresponding LOWN budget and include existing off-grid settings. Table~\ref{tab:competitor-settings} specifies the retention and KLASS thresholds selected for Table~\ref{tab:main}. The Avg. column in Table~\ref{tab:main} is the unweighted mean of $100\times$score$/$baseline score, except that CapArena uses $100+$score$-$baseline score.

\subsection{Judges}
MME reports perception plus cognition for LLaDA-V and perception alone for LaViDa, while CapArena reports a win--loss score. We score ScienceQA, ChartQA, MME, DocVQA, and GQA with task-specific rules. For LLaDA-V ScienceQA, we extract answers with the deterministic multiple-choice parser from lmms-eval v0.7.1. MathVista extracts final answers with GPT-4o-2024-08-06 before scoring, and CapArena is judged with the same model. MMBench uses rule matching followed by GPT-3.5-turbo-0125 when needed, scoring unmatched answers as incorrect, and LLaVA-Bench is judged with GPT-4o-2024-11-20. DeCapBench uses Qwen3.5-35B-A3B for its fact-based evaluation. All methods within each reported comparison use the same scorer.

\subsection{Baseline Settings}
\label{app:baselines}
Each competitor exposes a budget hyperparameter, the retention ratio $r$ for RedVTP and ZOO-Prune and the confidence and KL thresholds $(\tau,\kappa)$ for KLASS. We run each on a fixed grid whose ends keep the method within its working range, a 10\% grid of $r$ for RedVTP and ZOO-Prune on LLaDA-V together with existing off-grid settings, and a grid of threshold pairs for KLASS. Because the grid is discrete, a competitor's FLOPs cannot be set equal to LOWN's. For every benchmark we therefore select the cheapest grid setting whose FLOPs are at least those of the LOWN row it is compared with, and scores play no part in the selection. Every competitor cell in Tables~\ref{tab:main} and~\ref{tab:lavida} consequently spends at least as much compute as LOWN, so a competitor that scores below LOWN does so at equal or higher cost. Table~\ref{tab:competitor-settings} lists the selected settings on LLaDA-V, and Table~\ref{tab:lavida} shows the selected retention under each LaViDa cell.

Figure~\ref{fig:concept-c} matches scores instead of FLOPs. For each benchmark the target is LOWN's score, and each pruning method is drawn at its cheapest measured setting, up to 75\% retention, whose score is at least the target. When no setting up to 75\% reaches the target, the 75\% setting is drawn, so its bar is an upper bound on the speedup the method would have at LOWN's score. Only measured settings are used, without interpolation.

\begin{table}[ht]
\centering
\caption{Settings behind the competitor cells of Table~\ref{tab:main} (FLOPs matched) and Table~\ref{tab:wallclock-matched} (time matched). For RedVTP, $r$ is the visual-token retention percentage. For ZOO-Prune, the released code sets the budget to $r\%$ of a fixed 2,880 tokens rather than of each image's visual tokens, so the fraction actually retained varies with the image. For KLASS, $(\tau,\kappa)$ denotes its confidence and KL thresholds. The second and fourth blocks use the budget of LOWN composed with RedVTP at $r=50\%$.}
\label{tab:competitor-settings}
\small
\setlength{\tabcolsep}{3.2pt}
\resizebox{\linewidth}{!}{%
\begin{tabular}{@{}lcccccccccc@{}}
\toprule
 & MathVista & ScienceQA & ChartQA & MME & MMBench & DocVQA & CapArena & LLaVA-B. & GQA & DeCapB. \\
\midrule
\multicolumn{11}{@{}l}{\emph{FLOPs matched to LOWN}} \\
RedVTP $r$ & 30 & 20 & 60 & 20 & 20 & 50 & 40 & 50 & 20 & 40 \\
ZOO-Prune $r$ & 30 & 20 & 80 & 30 & 20 & 60 & 50 & 60 & 20 & 50 \\
KLASS $(\tau, \kappa)$ & (0.2, 0.2) & (0.2, 1) & (0.2, 1) & (0.2, 0.3) & (0.2, 0.4) & (0.2, 0.1) & (0.2, 0.5) & (0.2, 1) & (0.2, 1) & (0.2, 0.4) \\
\midrule
\multicolumn{11}{@{}l}{\emph{FLOPs matched to LOWN + RedVTP ($r=50\%$)}} \\
RedVTP $r$ & 20 & 10 & 30 & 10 & 20 & 25 & 20 & 25 & 10 & 20 \\
ZOO-Prune $r$ & 11.12 & 11.12 & 50 & 20 & 11.12 & 33.34 & 30 & 33.34 & 11.12 & 30 \\
KLASS + RedVTP $(\tau, \kappa)$ & (0.2, 0.1) & (0.2, 0.4) & (0.1, 0.2) & (0.2, 0.3) & (0.2, 0.4) & (0.2, 0.1) & (0.2, 0.5) & (0.2, 1) & (0.2, 1) & (0.2, 0.4) \\
\midrule
\multicolumn{11}{@{}l}{\emph{Time matched to LOWN}} \\
RedVTP $r$ & 30 & 10 & 50 & 10 & 10 & 50 & 50 & 60 & 10 & 50 \\
ZOO-Prune $r$ & 30 & 11.12 & 75 & 30 & 22.23 & 60 & 60 & 70 & 22.23 & 60 \\
KLASS $(\tau, \kappa)$ & (0.3, 0.1) & (0.2, 0.4) & (0.2, 0.5) & (0.2, 0.3) & (0.2, 0.1) & (0.3, 0.1) & (0.2, 0.5) & (0.2, 0.5) & (0.2, 0.1) & (0.2, 0.4) \\
\midrule
\multicolumn{11}{@{}l}{\emph{Time matched to LOWN + RedVTP ($r=50\%$)}} \\
RedVTP $r$ & 10 & 10 & 40 & 10 & 10 & 30 & 25 & 40 & 10 & 25 \\
ZOO-Prune $r$ & 11.12 & 11.12 & 70 & 30 & 22.23 & 50 & 33.34 & 40 & 22.23 & 33.34 \\
KLASS + RedVTP $(\tau, \kappa)$ & (0.4, 0.1) & (0.5, 0.1) & (0.2, 0.1) & (0.3, 0.1) & (0.5, 0.1) & (0.5, 0.1) & (0.2, 0.1) & (0.2, 0.1) & (0.3, 0.1) & (0.2, 0.1) \\
\bottomrule
\end{tabular}}
\end{table}

\subsection{Video Evaluation}
\label{app:video}
We evaluate the 500-question EgoSchema subset and the 502-question MLVU test set. We use VidLaDA-8B with 32 frames per video, sampled at one frame per second and uniformly subsampled when needed. Each frame yields 210 visual tokens after spatial pooling. We adopt a controlled decoding protocol with a 128-token response, 128 baseline diffusion steps, and block length 32, following the efficiency-ablation setup of \citet{he2026vidlada}. Decoding is greedy, with zero classifier-free guidance and low-confidence remasking. The baseline's parallel decoding is disabled, and LOWN uses adaptive acceptance within each block. We ask the model to reason inside \texttt{<think>} and \texttt{</think>} before answering with an option letter. This protocol differs from VidLaDA's released 64-frame, two-token evaluation.

The two methods are evaluated on identical questions and frames. We score only the answer span after the closing tag, using the released task parsers. EgoSchema reports accuracy. MLVU reports the mean accuracy over its nine task types (M-Avg). For each benchmark, the FLOPs ratio divides LOWN's summed decoding FLOPs by the baseline's sum over the same questions. It includes the decoder and LM head and excludes the vision encoder.

\FloatBarrier
\Needspace{6\baselineskip}
\section{Wall-Clock Measurements}
\label{app:wallclock}

\subsection{Generation Time}
All LLaDA-V timing experiments were run on the same server with NVIDIA H200 GPUs, with one experiment per GPU at a time. We measure per-example generation time at batch size 1 with a host clock around the model's generation call. CUDA is synchronized before the timer starts and after generation returns. The interval includes decoding, multimodal preparation, and vision encoding performed within the call. Model loading and evaluation scoring are excluded. Each task mean averages all per-example times from one recorded execution of the full evaluation set. FLOPs count the decoder, LM head, and method-specific matrix multiplications, so their scope differs from the timing interval. The cache comparison in Table~\ref{tab:fastdllm} includes 1,000 MathVista examples, 60 LLaVA-Bench questions, and 400 DeCapBench examples.

\subsection{Quality and Generation Time}
Table~\ref{tab:wallclock-matched} repeats the comparison of Table~\ref{tab:main} with each competitor selected by measured time. For each competitor, among its measured retentions (down to the floors of Table~\ref{tab:main}) and KLASS thresholds, we take the setting with the shortest mean generation time that is at least LOWN's, so no competitor is given less time than LOWN. When even the fastest setting is slower than LOWN, that setting is used. Accuracy plays no part in the selection. Cells within a few percent of LOWN's time should be read as equal in cost. With every competitor given at least LOWN's generation time on every benchmark, LOWN keeps the highest average in both blocks, 99.6\% and 98.3\% of LLaDA-V against at most 94.6\% and 92.8\% for the competitors, and scores highest on seven and eight of the ten benchmarks, respectively.

\begin{table}[ht]
\centering
\caption{\textbf{Performance at matched generation time on LLaDA-V}. The upper block compares standalone methods at LOWN's generation time, and the lower block at that of LOWN with RedVTP ($r{=}50\%$). Competitors run on a discrete hyperparameter grid, so each cell uses the setting with the shortest measured time at or above LOWN's (Table~\ref{tab:competitor-settings}), and LOWN gives the highest normalized average in both blocks. Cells show scores with mean generation time relative to LLaDA-V below. Best scores are bold, second best underlined, and lowest times bold.}
\label{tab:wallclock-matched}
\scriptsize
\setlength{\tabcolsep}{2pt}
\renewcommand{\arraystretch}{1.0}
\setlength{\aboverulesep}{0pt}\setlength{\belowrulesep}{0pt}\setlength{\extrarowheight}{1pt}
\resizebox{\linewidth}{!}{%
\begin{tabular}{l cccccccccc|c}
\toprule
 & Math & Science & Chart & & MM & Doc & Cap & LLaVA- & & DeCap & \\
\multirow{-2}{*}{Method} & Vista & QA & QA & \multirow{-2}{*}{MME} & Bench & VQA & Arena & Bench & \multirow{-2}{*}{GQA} & Bench & \multirow{-2}{*}{Avg.} \\
\midrule
LLaDA-V \textit{(Vanilla)} & 59.10 & 76.50 & 78.16 & 1719.49 & 70.88 & 83.91 & $-$88.67 & 58.20 & 53.31 & 33.18 & 100.0\% \\
\midrule
\multicolumn{11}{>{\columncolor{block-grey}}c|}{\textit{Time matched to LOWN}} & \cellcolor{block-grey} \\
\midrule
\rowcolor{white} & 51.70 & 65.49 & 72.20 & 1275.08 & 40.12 & \textbf{83.39} & \underline{$-$88.50} & \textbf{57.60} & 48.05 & \underline{33.88} & 88.7\% \\
\multirow{-2}{*}{RedVTP} & \fl{0.394} & \fl{0.374} & \fl{0.674} & \fl{0.300} & \fl{0.366} & \fl{0.566} & \fl{0.507} & \fl{0.612} & \fl{0.318} & \fl{0.504} & \fl{0.461} \\
& 53.70 & 72.48 & \underline{76.00} & 1582.41 & \underline{68.73} & 78.73 & $-$94.67 & 56.70 & \underline{50.50} & 31.15 & \underline{94.6\%} \\
\multirow{-2}{*}{ZOO-Prune} & \fl{0.417} & \fl{0.331} & \fl{0.690} & \fl{0.352} & \fl{0.393} & \fl{0.548} & \fl{0.490} & \fl{0.587} & \fl{0.301} & \fl{0.488} & \fl{0.460} \\
& \textbf{56.90} & \underline{76.05} & 73.56 & \underline{1641.99} & 68.56 & 76.67 & $-$96.83 & 42.20 & 49.55 & 30.18 & 92.2\% \\
\multirow{-2}{*}{KLASS} & \fl{0.493} & \fl{0.337} & \fl{0.676} & \fl{0.309} & \fl{0.387} & \fl{0.560} & \fl{0.450} & \fl{0.573} & \fl{0.284} & \fl{0.479} & \fl{0.455} \\
\rowcolor{line-blue} & \underline{56.00} & \textbf{78.83} & \textbf{78.36} & \textbf{1655.08} & \textbf{69.93} & \underline{82.31} & \textbf{$-$85.67} & \underline{57.40} & \textbf{53.24} & \textbf{34.43} & \textbf{99.6\%} \\
\rowcolor{line-blue} \multirow{-2}{*}{LOWN (Ours)} & \flb{0.380} & \flb{0.331} & \flb{0.666} & \flb{0.295} & \flb{0.362} & \flb{0.537} & \flb{0.441} & \flb{0.532} & \flb{0.268} & \flb{0.446} & \flb{0.426} \\
\midrule
\multicolumn{11}{>{\columncolor{block-grey}}c|}{\textit{Time matched to LOWN (Ours) + RedVTP ($r{=}50\%$)}} & \cellcolor{block-grey} \\
\midrule
\rowcolor{white} & 41.70 & 65.49 & 70.80 & 1275.08 & 40.12 & \underline{82.14} & \underline{$-$89.33} & \textbf{57.10} & 48.05 & \underline{32.28} & 86.0\% \\
\multirow{-2}{*}{RedVTP} & \fl{0.279} & \fl{0.374} & \fl{0.635} & \fl{0.300} & \fl{0.366} & \fl{0.408} & \fl{0.299} & \fl{0.436} & \fl{0.318} & \fl{0.296} & \fl{0.371} \\
& 46.10 & 72.48 & \textbf{74.92} & 1582.41 & \underline{68.73} & 75.97 & $-$96.17 & 52.90 & 50.50 & 29.56 & 91.5\% \\
\multirow{-2}{*}{ZOO-Prune} & \fl{0.264} & \fl{0.331} & \fl{0.648} & \fl{0.352} & \fl{0.393} & \fl{0.453} & \fl{0.301} & \fl{0.357} & \fl{0.301} & \fl{0.299} & \fl{0.370} \\
KLASS & \underline{54.70} & \underline{74.67} & 70.64 & \underline{1606.29} & 68.38 & 82.07 & $-$96.00 & 47.20 & \underline{50.74} & 30.13 & \underline{92.8\%} \\
\quad + RedVTP \textit{($r{=}50\%$)} & \fl{0.327} & \fl{0.370} & \fl{0.657} & \fl{0.306} & \fl{0.364} & \fl{0.418} & \fl{0.296} & \fl{0.371} & \fl{0.261} & \fl{0.299} & \fl{0.367} \\
\rowcolor{line-blue} LOWN (Ours) & \textbf{54.90} & \textbf{75.56} & \underline{74.88} & \textbf{1709.75} & \textbf{69.67} & \textbf{82.24} & \textbf{$-$85.17} & \underline{55.00} & \textbf{53.42} & \textbf{33.61} & \textbf{98.3\%} \\
\rowcolor{line-blue} \quad + RedVTP \textit{($r{=}50\%$)} & \flb{0.262} & \flb{0.319} & \flb{0.577} & \flb{0.276} & \flb{0.333} & \flb{0.398} & \flb{0.284} & \flb{0.354} & \flb{0.256} & \flb{0.278} & \flb{0.334} \\
\bottomrule
\end{tabular}
}
\end{table}

\FloatBarrier
\subsection{Cache Configuration and Composition}
Fast-dLLM uses a block spanning the configured output, prefix refresh every 32 forwards, and its factor-based token selection with factor 1. The combined implementation retains these cache settings, maintains separate image-conditioned and TEXT-ONLY prefix caches, and takes the union of Fast-dLLM's selected tokens and LOWN's accepted tokens.

\clearpage
\section{Additional Results}
\label{app:tables}

\subsection{Fixed-Retention Pruning Compositions}
Table~\ref{tab:pruning-relative} measures the effect of adding LOWN to the unpruned model and to each fixed pruner, using the corresponding configuration without LOWN as the reference. LOWN retains much of its standalone compute reduction when applied on top of pruning, with average quality close to that of each reference configuration.

\begin{table}[ht]
\centering
\caption{Relative gains from adding LOWN on LLaDA-V and LaViDa. Each row is measured against the listed configuration without LOWN. FLOPs ratio is the mean of the per-benchmark FLOPs ratios, and reduction is one minus this mean. Quality is the mean normalized index with the reference set to 100.}
\label{tab:pruning-relative}
\small
\setlength{\tabcolsep}{5pt}
\resizebox{0.75\linewidth}{!}{%
\begin{tabular}{@{}lrrr@{}}
\toprule
Configuration before LOWN & FLOPs ratio $\downarrow$ & Reduction $\uparrow$ & Quality index $\uparrow$ \\
\midrule
\multicolumn{4}{@{}l}{\emph{LLaDA-V}} \\
LLaDA-V (unpruned) & 0.353 & 64.7\% & 99.6 \\
RedVTP (50\% retention) & 0.424 & 57.6\% & 100.9 \\
ZOO-Prune (33\% retention) & 0.394 & 60.6\% & 98.7 \\
\midrule
\multicolumn{4}{@{}l}{\emph{LaViDa}} \\
LaViDa (unpruned) & 0.174 & 82.6\% & 99.7 \\
RedVTP (75\% retention) & 0.227 & 77.3\% & 99.5 \\
ZOO-Prune (33\% retention) & 0.278 & 72.2\% & 99.8 \\
\bottomrule
\end{tabular}}
\end{table}

\subsection{Fewer Denoising Steps}
\label{app:fewer-steps}
Table~\ref{tab:fewer-steps} runs LLaDA-V with $T/2$ and $T/4$ denoising steps on the CapArena, LLaVA-Bench, and DeCapBench inputs of Table~\ref{tab:main}. LOWN's FLOPs lie between the two settings, and LOWN scores above both on all three benchmarks, including $T/2$ at 0.500$\times$.

\begin{table}[h]
\centering
\caption{LLaDA-V with fewer denoising steps on CapArena, LLaVA-Bench, and DeCapBench. $T/2$ and $T/4$ run half and a quarter of the baseline's $T$ steps on the inputs of Table~\ref{tab:main}. Scores use the judges of Table~\ref{tab:main}, and FLOPs are relative to LLaDA-V. DeCapBench averages the captions its official evaluator judges valid, 399, 398, 392, and 399 of 400 for the four rows.}
\label{tab:fewer-steps}
\small
\setlength{\tabcolsep}{4pt}
\begin{tabular}{@{}lcccccc@{}}
\toprule
 & \multicolumn{2}{c}{CapArena} & \multicolumn{2}{c}{LLaVA-Bench} & \multicolumn{2}{c}{DeCapBench} \\
\cmidrule(lr){2-3} \cmidrule(lr){4-5} \cmidrule(l){6-7}
Method & Score $\uparrow$ & FLOPs $\downarrow$ & Score $\uparrow$ & FLOPs $\downarrow$ & Score $\uparrow$ & FLOPs $\downarrow$ \\
\midrule
LLaDA-V ($T$ steps) & $-$88.67 & 1.000 & 58.20 & 1.000 & 33.18 & 1.000 \\
LLaDA-V ($T/2$) & $-$92.33 & 0.500 & 45.40 & 0.500 & 32.17 & 0.500 \\
LLaDA-V ($T/4$) & $-$99.00 & 0.250 & 26.20 & 0.250 & 25.84 & 0.250 \\
\rowcolor{line-blue} LOWN (Ours) & $-$85.67 & 0.388 & 57.40 & 0.454 & 34.43 & 0.391 \\
\bottomrule
\end{tabular}
\end{table}

\clearpage
\subsection{LaViDa}
\label{app:lavida}
\begin{figure}[h]
    \centering
    \includegraphics[width=\linewidth]{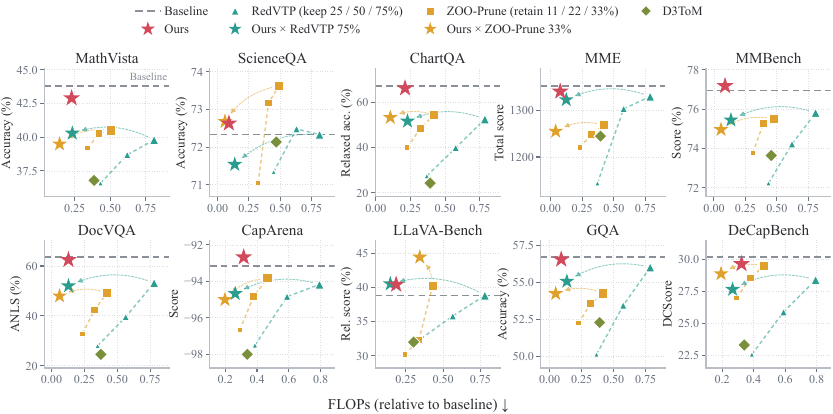}
    \caption{\textbf{Performance and efficiency on LaViDa, all ten benchmarks.} The markers follow Figure~\ref{fig:pareto-lavida}.}
    \label{fig:pareto-lavida-full}
\end{figure}

Figure~\ref{fig:pareto-lavida-full} shows all ten benchmarks, and Table~\ref{tab:lavida} gives the FLOPs-matched comparison in the format of Table~\ref{tab:main}.

\begin{table}[h]
\centering
\caption{\textbf{Performance at matched decoding FLOPs on LaViDa}, in the format of Table~\ref{tab:main}. Methods are compared at LOWN's budget. Each cell shows the benchmark score, with the selected retention $r$ and FLOPs relative to the LaViDa baseline below. Competitors run on a discrete grid of retention ratios, so each uses its cheapest measured setting whose FLOPs are at least LOWN's (Appendix~\ref{app:baselines}). D$^3$ToM uses constant 10\% retention on MathVista, LLaVA-Bench, CapArena, and DeCapBench and its adaptive schedule on the remaining tasks. MME reports the perception score. Best scores are bold, second best underlined, and lowest FLOPs bold.}
\label{tab:lavida}
\scriptsize
\setlength{\tabcolsep}{2pt}
\renewcommand{\arraystretch}{1.0}
\setlength{\aboverulesep}{0pt}\setlength{\belowrulesep}{0pt}\setlength{\extrarowheight}{1pt}
\resizebox{\linewidth}{!}{%
\begin{tabular}{l cccccccccc|c}
\toprule
 & Math & Science & Chart & & MM & Doc & Cap & LLaVA- & & DeCap & \\
\multirow{-2}{*}{Method} & Vista & QA & QA & \multirow{-2}{*}{MME} & Bench & VQA & Arena & Bench & \multirow{-2}{*}{GQA} & Bench & \multirow{-2}{*}{Avg.} \\
\midrule
LaViDa \textit{(Vanilla)} & 43.80 & 72.34 & 67.16 & 1353.03 & 76.95 & 63.54 & $-$93.17 & 38.80 & 56.69 & 30.18 & 100.0\% \\
\midrule
\multicolumn{11}{>{\columncolor{block-grey}}c|}{\textit{FLOPs matched to LOWN}} & \cellcolor{block-grey} \\
\midrule
\rowcolor{white}  & 36.60 & 71.34 & 20.40 & 987.28 & 72.23 & 18.55 & $-$99.33 & 28.60 & 50.14 & 22.58 & 73.9\% \\
\multirow{-2}{*}{RedVTP} & \ret{25}{0.432} & \ret{25}{0.448} & \ret{10}{0.244} & \ret{10}{0.248} & \ret{25}{0.435} & \ret{10}{0.225} & \ret{20}{0.347} & \ret{10}{0.226} & \ret{25}{0.367} & \ret{25}{0.389} & \fl{0.336} \\
 & \underline{39.20} & 71.05 & \underline{39.56} & 1220.76 & \underline{73.76} & \underline{30.62} & \underline{$-$95.67} & 30.20 & 52.23 & \underline{28.55} & \underline{84.3\%} \\
\multirow{-2}{*}{ZOO-Prune} & \ret{11}{0.341} & \ret{11}{0.324} & \ret{10}{0.217} & \ret{11}{0.232} & \ret{11}{0.309} & \ret{10}{0.225} & \ret{15}{0.321} & \ret{11}{0.245} & \ret{11}{0.227} & \ret{22}{0.379} & \fl{0.282} \\
 & 36.80 & \underline{72.14} & 24.24 & \underline{1244.92} & 73.64 & 24.56 & $-$98.00 & \underline{32.00} & \underline{52.28} & 23.31 & 79.3\% \\
\multirow{-2}{*}{D$^3$ToM} & \fl{0.387} & \fl{0.469} & \fl{0.392} & \fl{0.396} & \fl{0.456} & \fl{0.376} & \fl{0.339} & \fl{0.302} & \fl{0.391} & \fl{0.341} & \fl{0.385} \\
\rowcolor{line-blue} & \textbf{42.90} & \textbf{72.63} & \textbf{66.16} & \textbf{1341.15} & \textbf{77.18} & \textbf{62.49} & \textbf{$-$92.67} & \textbf{40.40} & \textbf{56.54} & \textbf{29.65} & \textbf{99.7\%} \\
\rowcolor{line-blue} \multirow{-2}{*}{LOWN (Ours)} & \flb{0.229} & \flb{0.091} & \flb{0.211} & \flb{0.078} & \flb{0.088} & \flb{0.130} & \flb{0.316} & \flb{0.185} & \flb{0.091} & \flb{0.323} & \flb{0.174} \\
\bottomrule
\end{tabular}
}
\end{table}

\subsection{Qualitative Samples}
\begin{figure}[t]
    \centering
    \includegraphics[width=0.72\linewidth]{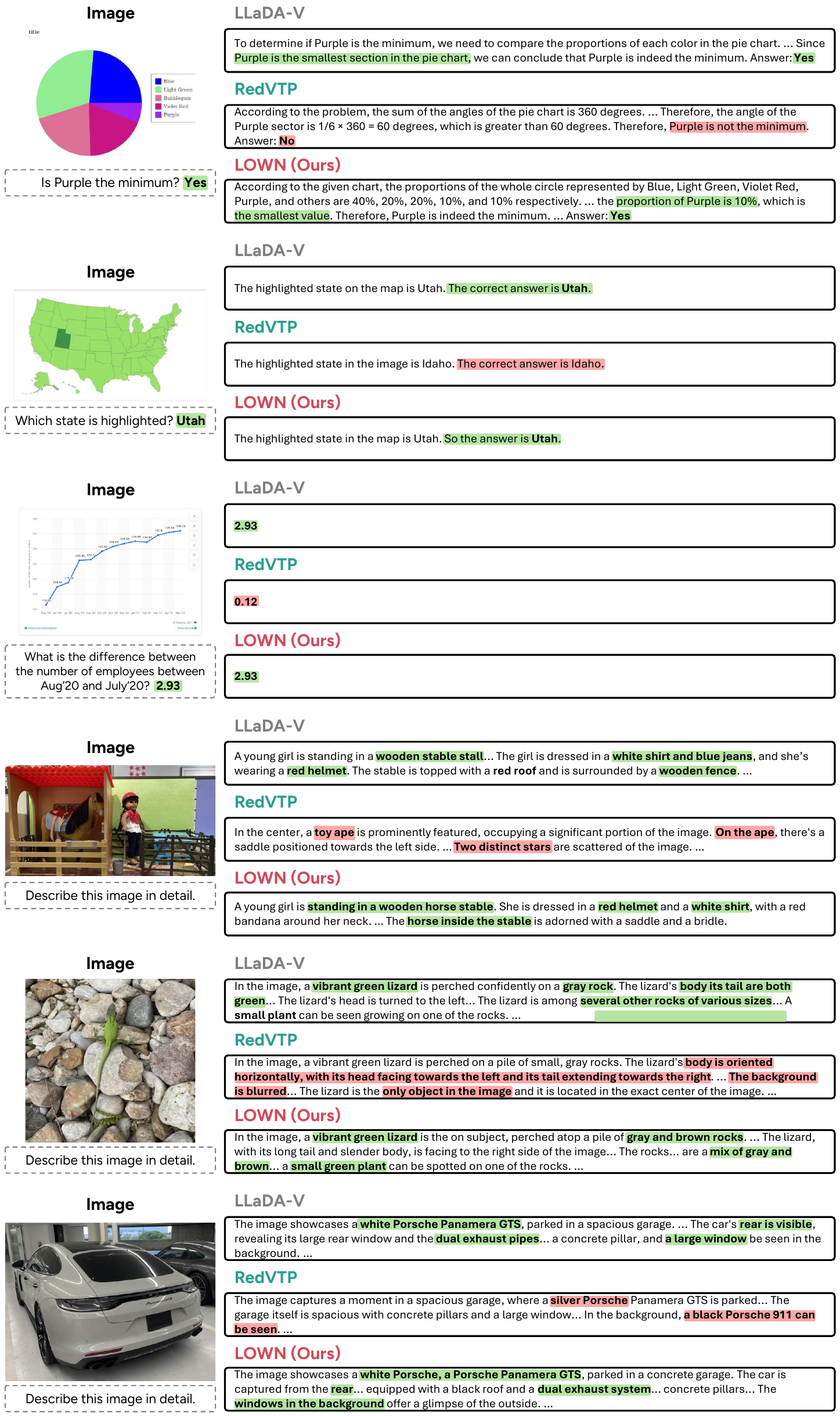}
    \caption{\textbf{Additional qualitative comparisons with visual token pruning (RedVTP).} The upper three examples are short-answer questions, and the lower three are open-ended captioning. Green marks correct answers and content consistent with the image, and red marks incorrect answers and content inconsistent with the image.}
    \label{fig:LOWN_vs_Exsitings_appx}
\end{figure}

\begin{figure}[t]
    \centering
    \includegraphics[width=0.85\linewidth]{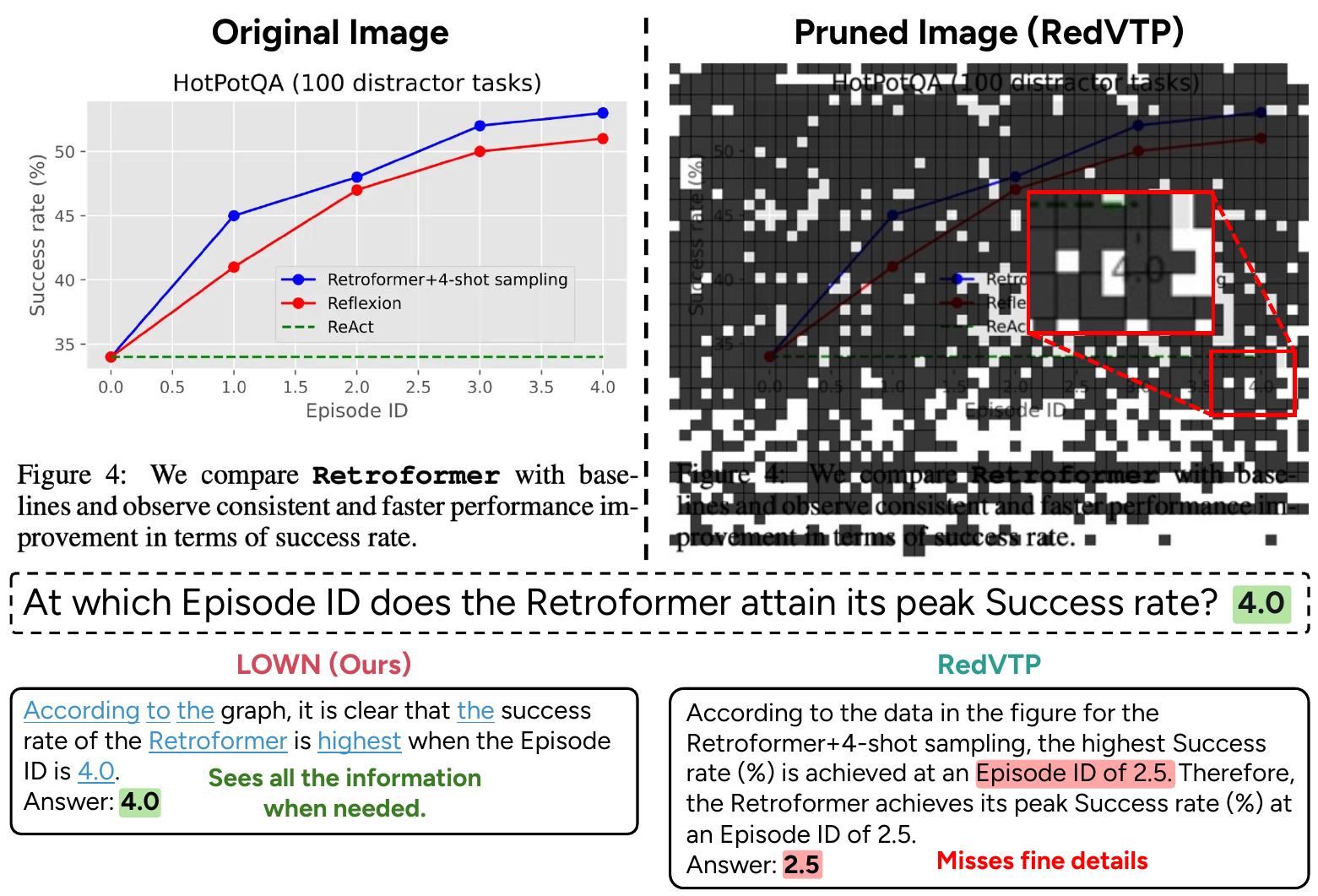}
    \caption{\textbf{Qualitative comparison with visual token pruning (RedVTP) and its pruned image.} The upper right darkens the visual tokens removed by RedVTP, and the inset enlarges the region around the x-axis tick 4.0. In LOWN's output, blue underlines mark tokens unmasked on FULL forward steps, while the remaining tokens are unmasked on TEXT-ONLY forward steps.}
    \label{fig:LOWN_vs_Pruned}
\end{figure}

Figures~\ref{fig:LOWN_vs_Exsitings_appx} and~\ref{fig:LOWN_vs_Pruned} extend the comparison of Figure~\ref{fig:LOWN_vs_Exsitings} to more examples. In Figure~\ref{fig:LOWN_vs_Exsitings_appx}, RedVTP answers the short-answer questions incorrectly and, in open-ended captioning, describes objects and attributes that do not match the image. LOWN gives the same answers as LLaDA-V and describes the same main objects. In Figure~\ref{fig:LOWN_vs_Pruned}, RedVTP processes the image at every step, but after pruning it processes only the visual tokens it retains, which exclude the x-axis tick 4.0 that answers the question. Its response is fluent but ends with the incorrect answer 2.5. LOWN processes the image on fewer steps, but each FULL forward covers the complete image.
\clearpage

\clearpage

\subsection{Limitation Example}
\label{app:limitation}
\begin{figure}[h]
    \centering
    \includegraphics[width=0.85\linewidth]{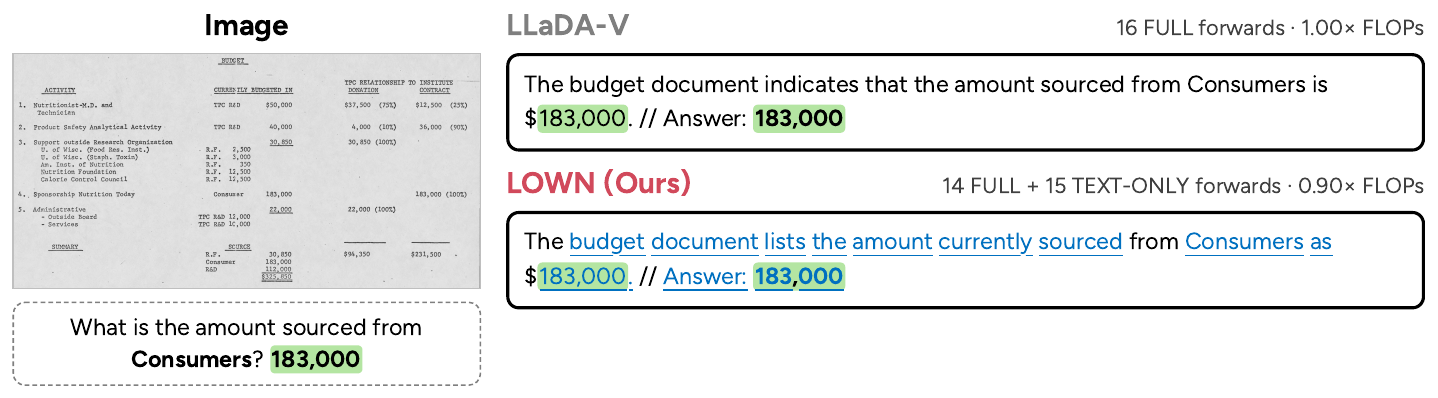}
    \caption{\textbf{A limitation example on DocVQA.} In LOWN's output, blue underlines mark tokens unmasked on FULL forward steps, while the remaining tokens are unmasked on TEXT-ONLY forward steps. FLOPs are relative to LLaDA-V on this input.}
    \label{fig:limitation}
\end{figure}

Figure~\ref{fig:limitation} shows the DocVQA input on which LOWN saves the least. The question asks for a number in a table, and LOWN runs a FULL forward on 14 of its 15 steps. LOWN answers correctly but costs 0.90$\times$ the baseline FLOPs, against 0.45$\times$ over the whole benchmark (Table~\ref{tab:main}).

\FloatBarrier
\Needspace{6\baselineskip}
\section{Additional Ablations and Analysis}
\label{app:ablation}

\subsection{Order of FULL-Step Unmasking}
\label{app:fullorder}
Table~\ref{tab:fullorder} separates the two parts of adaptive acceptance in Table~\ref{tab:components}. Disagreement-first order removes 14--35\% of the cost of visual routing alone while scoring higher on seven of the ten benchmarks, and whole-prefix acceptance removes a further 2--19\%.

\begin{table}[h]
\centering
\caption{Table~\ref{tab:components} with adaptive acceptance split into disagreement-first order on FULL steps and whole-prefix acceptance on TEXT-ONLY steps (Section~\ref{sec:step}), on LLaDA-V. Each setting reports scores and, below, relative FLOPs with the reduction from the preceding setting in parentheses.}
\label{tab:fullorder}
\scriptsize
\setlength{\tabcolsep}{2pt}
\renewcommand{\arraystretch}{1.0}
\resizebox{\linewidth}{!}{%
\begin{tabular}{@{}ccc cccccccccc@{}}
\toprule
 & \multicolumn{2}{c}{Adaptive acceptance} & Math & Science & Chart & & MM & Doc & Cap & LLaVA- & & DeCap \\
\cmidrule(lr){2-3}
\multirow{-2}{*}{Visual routing} & Disagreement & Whole prefix & Vista & QA & QA & \multirow{-2}{*}{MME} & Bench & VQA & Arena & Bench & \multirow{-2}{*}{GQA} & Bench \\
\midrule
\xmark & \xmark & \xmark & 59.10 & 76.50 & 78.16 & 1719.5 & 70.88 & 83.91 & $-$88.67 & 58.20 & 53.31 & 33.18 \\
\addlinespace[1.5pt]\hdashline[0.5pt/1.5pt]\addlinespace[1.5pt]
 & & & 57.60 & 77.39 & 77.76 & 1624.6 & 68.30 & 83.50 & $-$86.83 & 57.90 & 52.53 & 33.02 \\
\multirow{-2}{*}{\cmark} & \multirow{-2}{*}{\xmark} & \multirow{-2}{*}{\xmark} & 0.5040 \pc{49.6} & 0.3554 \pc{64.5} & 0.7330 \pc{26.7} & 0.3383 \pc{66.2} & 0.3759 \pc{62.4} & 0.6743 \pc{32.6} & 0.6833 \pc{31.7} & 0.7002 \pc{30.0} & 0.3203 \pc{68.0} & 0.6840 \pc{31.6} \\
\addlinespace[1.5pt]\hdashline[0.5pt/1.5pt]\addlinespace[1.5pt]
 & & & 57.10 & 78.19 & 78.32 & 1688.5 & 69.85 & 82.33 & $-$81.17 & 57.50 & 53.43 & 34.71 \\
\multirow{-2}{*}{\cmark} & \multirow{-2}{*}{\cmark} & \multirow{-2}{*}{\xmark} & 0.3910 \pc{22.4} & 0.3006 \pc{15.4} & 0.5891 \pc{19.6} & 0.2755 \pc{18.6} & 0.3236 \pc{13.9} & 0.4789 \pc{29.0} & 0.4471 \pc{34.6} & 0.5067 \pc{27.6} & 0.2469 \pc{22.9} & 0.4450 \pc{34.9} \\
\addlinespace[1.5pt]\hdashline[0.5pt/1.5pt]\addlinespace[1.5pt]
 & & & 56.00 & 78.83 & 78.36 & 1655.1 & 69.93 & 82.31 & $-$85.67 & 57.40 & 53.24 & 34.43 \\
\multirow{-2}{*}{\cmark} & \multirow{-2}{*}{\cmark} & \multirow{-2}{*}{\cmark} & 0.3252 \pc{16.8} & 0.2428 \pc{19.2} & 0.5759 \pc{2.2} & 0.2307 \pc{16.3} & 0.2701 \pc{16.5} & 0.4517 \pc{5.7} & 0.3885 \pc{13.1} & 0.4543 \pc{10.3} & 0.2045 \pc{17.2} & 0.3909 \pc{12.2} \\
\bottomrule
\end{tabular}}
\end{table}

\subsection{Certificate Reliability}
\label{app:analysis}
Coverage, $\epsilon(\Delta)$, and $R_{\mathrm{ref}}$ are computed on the logged baseline trajectories of Appendix~\ref{app:replay-protocol}, and the acceptance error and anchor age on LOWN's own decoding of the same samples (Tables~\ref{tab:cert} and~\ref{tab:anchor-age}).

At anchor age one the certificate certifies 43--72\% of the eligible positions, $\epsilon(\Delta)$ rises from age one to four on every benchmark, and the acceptance error stays below $\epsilon(1)$ on every benchmark. On DocVQA, LOWN costs less than the oracle schedule because adaptive acceptance ends the trajectory early.

\begin{table}[!h]
\centering
\caption{Certificate reliability on LLaDA-V. Eligible positions are the masked positions that were already masked $\Delta$ steps earlier. Coverage is the share of eligible positions whose TEXT-ONLY prediction agrees with a one-step-old anchor. $\epsilon(\Delta)$ is the share of agreeing positions at which a fresh FULL forward predicts a different token when the anchor is $\Delta$ steps old. $R_{\mathrm{ref}}=f^{*}+(1-f^{*})\rho+1/T$ is the cost of the oracle schedule of Section~\ref{sec:ablation}, which runs a FULL forward only on the fraction $f^{*}$ of steps where the image changes an unmasked decision and pays $\rho$ elsewhere, plus one initial FULL forward. For ChartQA, whose baseline stops at the end-of-sequence token after 7.0 of its 16 steps on average, $f^{*}$ and $T$ are taken over the steps the baseline actually runs ($f^{*}=32.3\%$). Acceptance error is the share of the tokens LOWN unmasks on TEXT-ONLY steps that a fresh FULL forward on the state they were accepted from would predict differently, and $R_{\mathrm{LOWN}}$ is LOWN's measured cost. Costs are relative to the baseline and rates are in percent.}
\label{tab:cert}
\footnotesize
\setlength{\tabcolsep}{4pt}
\begin{tabular}{@{}lrrrrrrrr@{}}
\toprule
Benchmark & Coverage & $\epsilon(1)$ & $\epsilon(2)$ & $\epsilon(4)$ & $\epsilon(8)$ & $R_{\mathrm{ref}}$ & Accept.\ error & $R_{\mathrm{LOWN}}$ \\
\midrule
MathVista & 58.6 & 7.3 & 9.5 & 12.2 & 15.4 & 0.284 & 2.2 & 0.325 \\
ScienceQA & 63.5 & 6.4 & 8.3 & 9.2 & 8.8 & 0.218 & 4.7 & 0.243 \\
ChartQA & 47.0 & 10.6 & 14.0 & 16.6 & 15.2 & 0.490 & 4.2 & 0.576 \\
MME & 63.8 & 5.9 & 7.6 & 10.0 & 13.6 & 0.193 & 3.9 & 0.231 \\
MMBench & 65.3 & 6.3 & 7.3 & 7.5 & 6.9 & 0.217 & 3.9 & 0.270 \\
DocVQA & 43.4 & 11.0 & 15.5 & 20.2 & 25.3 & 0.517 & 3.0 & 0.452 \\
CapArena & 72.1 & 5.2 & 6.8 & 8.7 & 10.8 & 0.374 & 3.5 & 0.388 \\
LLaVA-Bench & 54.0 & 9.5 & 12.6 & 16.1 & 19.5 & 0.430 & 4.2 & 0.454 \\
GQA & 64.7 & 4.1 & 5.4 & 7.0 & 11.1 & 0.200 & 2.3 & 0.205 \\
DeCapBench & 70.6 & 5.5 & 7.3 & 9.2 & 11.6 & 0.380 & 3.4 & 0.391 \\
\midrule
Mean & & & & & & 0.330 & & 0.353 \\
\bottomrule
\end{tabular}
\end{table}

\begin{table}[!h]
\centering
\caption{Anchor age of the tokens LOWN accepts on TEXT-ONLY steps, in LOWN's own runs on LLaDA-V. Age is the number of LOWN steps since the FULL forward that produced the anchor. Shares are in percent, and the last row pools the accepted tokens of all benchmarks.}
\label{tab:anchor-age}
\footnotesize
\setlength{\tabcolsep}{5pt}
\begin{tabular}{@{}lrrr@{}}
\toprule
Benchmark & Age 1 & Age $\le 2$ & Max age \\
\midrule
MathVista & 93.8 & 99.5 & 5 \\
ScienceQA & 92.9 & 99.4 & 3 \\
ChartQA & 94.9 & 99.9 & 3 \\
MME & 94.2 & 99.5 & 4 \\
MMBench & 91.8 & 99.4 & 3 \\
DocVQA & 96.7 & 99.6 & 4 \\
CapArena & 95.4 & 99.8 & 3 \\
LLaVA-Bench & 93.9 & 99.7 & 4 \\
GQA & 94.8 & 99.5 & 4 \\
DeCapBench & 95.3 & 99.9 & 3 \\
\midrule
All & 94.8 & 99.7 & 5 \\
\bottomrule
\end{tabular}
\end{table}

\FloatBarrier
\end{document}